\documentclass[11pt]{article}

\usepackage[preprint]{acl}

\usepackage{times}
\usepackage{latexsym}
\usepackage{booktabs} 

\usepackage{amsmath,amsfonts,bm}

\def\eqref#1{equation~\ref{#1}}

\def\1{\bm{1}}

\DeclareMathAlphabet{\mathsfit}{\encodingdefault}{\sfdefault}{m}{sl}
\SetMathAlphabet{\mathsfit}{bold}{\encodingdefault}{\sfdefault}{bx}{n}

\DeclareMathOperator*{\argmax}{arg\,max}

\usepackage{amssymb}
\usepackage{amsmath}
\usepackage{bbm}
\usepackage{hyperref}
\usepackage{url}

\usepackage{enumitem}
\usepackage{graphicx} 
\usepackage{algorithm}%
\usepackage{subcaption}
\usepackage{dsfont}
\usepackage{array}
\usepackage{algorithm}
\usepackage{algpseudocode}
\usepackage[dvipsnames]{xcolor}

\usepackage[T1]{fontenc}

\usepackage[utf8]{inputenc}

\usepackage{microtype}

\usepackage{inconsolata}

\usepackage{graphicx}

\title{Emergence and Localisation of Semantic Role Circuits in LLMs}

\author{Nura Aljaafari$^{1}$,~ Danilo S. Carvalho$^{3}$,~ Andr\'{e} Freitas$^{1,2,3}$ \\
  $^{1}$ Department of Computer Science, University of Manchester, United Kingdom\\
  $^{2}$ Idiap Research Institute, Switzerland\\
  $^{3}$ National Biomarker Centre, CRUK-MI, Univ. of Manchester, United Kingdom\\
  \texttt{\{firstname.lastname\}@$^{\dagger}$manchester.ac.uk}\\
}

\begin{document}
\maketitle
\begin{abstract}
Despite displaying semantic competence, large language models' internal mechanisms that ground abstract semantic structure remain insufficiently characterised. To investigate whether and how LLMs develop causally functional representations of semantic roles, we introduce a causal-temporal methodology combining contrastive minimal pairs, edge-attribution circuit discovery, and training-time tracking. Our analysis reveals that LLMs encode semantic roles through highly localised circuits (89–92\% attribution within $\leq$28 nodes) that emerge gradually via structural refinement rather than phase transitions. These circuits exhibit moderate cross-scale conservation (24–51\% component overlap) alongside high spectral similarity, with larger models reusing similar components while rewiring connections. These findings suggest that LLMs form compact, causally isolated mechanisms for abstract semantic structure that exhibit partial transfer across scales and architectures. 


\end{abstract}

\section{Introduction}\label{sec:intro}

\textbf{Do LLMs develop abstract, causally functional representations of semantic structure?}  
Large language models (LLMs) exhibit localised circuits for factual recall \citep{goldowskydill2023localizing, meng2022locating}, arithmetic \citep{conmy2023towards, Stolfo2023AMI}, and logical reasoning \citep{kim2025reasoning}. However, it remains unclear whether such mechanisms extend to the \emph{abstract relational semantic structure} that underlies natural language understanding. Current mechanistic studies often focus on specialised algorithmic behaviours in trained models (e.g., induction heads, copying, factual associations \citep{meng2022locating, gpt2_attention_saes}), leaving two central gaps in our understanding of semantic representations in LLMs.

\textbf{First, abstract semantic structure.} Semantic roles (e.g., \textsc{Agent}, \textsc{Theme}, \textsc{Instrument}) constitute a prevalent descriptive component of predicate–argument structure that generalises across surface forms and syntactic realisations in natural language \citep{fillmore1976frame}. Formally, a semantic role~$r$ associates a predicate~$p$ and argument position~$i$ with a thematic relation: for instance, in \emph{``The children played in the garden''}, the \textsc{Location} role links \emph{garden} to the playing event whether expressed as \emph{``in the garden''}, \emph{``at the garden''}, or \emph{``outside in the garden''} (see App.~\ref{app:linguistic-background} for a complete formalisation). Mechanistically, such predicate–argument binding requires integrating information across tokens and abstracting from surface cues, unlike behaviours explained by positional heuristics or lexical templates.  Whether LLMs implement this binding through \emph{causally functional circuits} however remains an open question.

\textbf{Second, temporal emergence.}  
Most circuit analyses examine only final checkpoints, obscuring \emph{when} semantic mechanisms arise, stabilise, and become computationally indispensable. Given that several behaviours emerge through sharp transitions (e.g., in-context learning, algorithmic generalisation \citep{wei2022emergent, nanda2023progress, he2024learning}), it is unclear whether semantic-role circuits likewise appear abruptly or gradually consolidate.  Understanding this timeline is important both for training-time interventions \citep{aljaafari2025trace2, cheng2024potentiallimitationsllmscapturing} and for explaining how syntactic and semantic abstractions co-develop.
\begin{figure*}
    \centering
    \includegraphics[width=\linewidth]{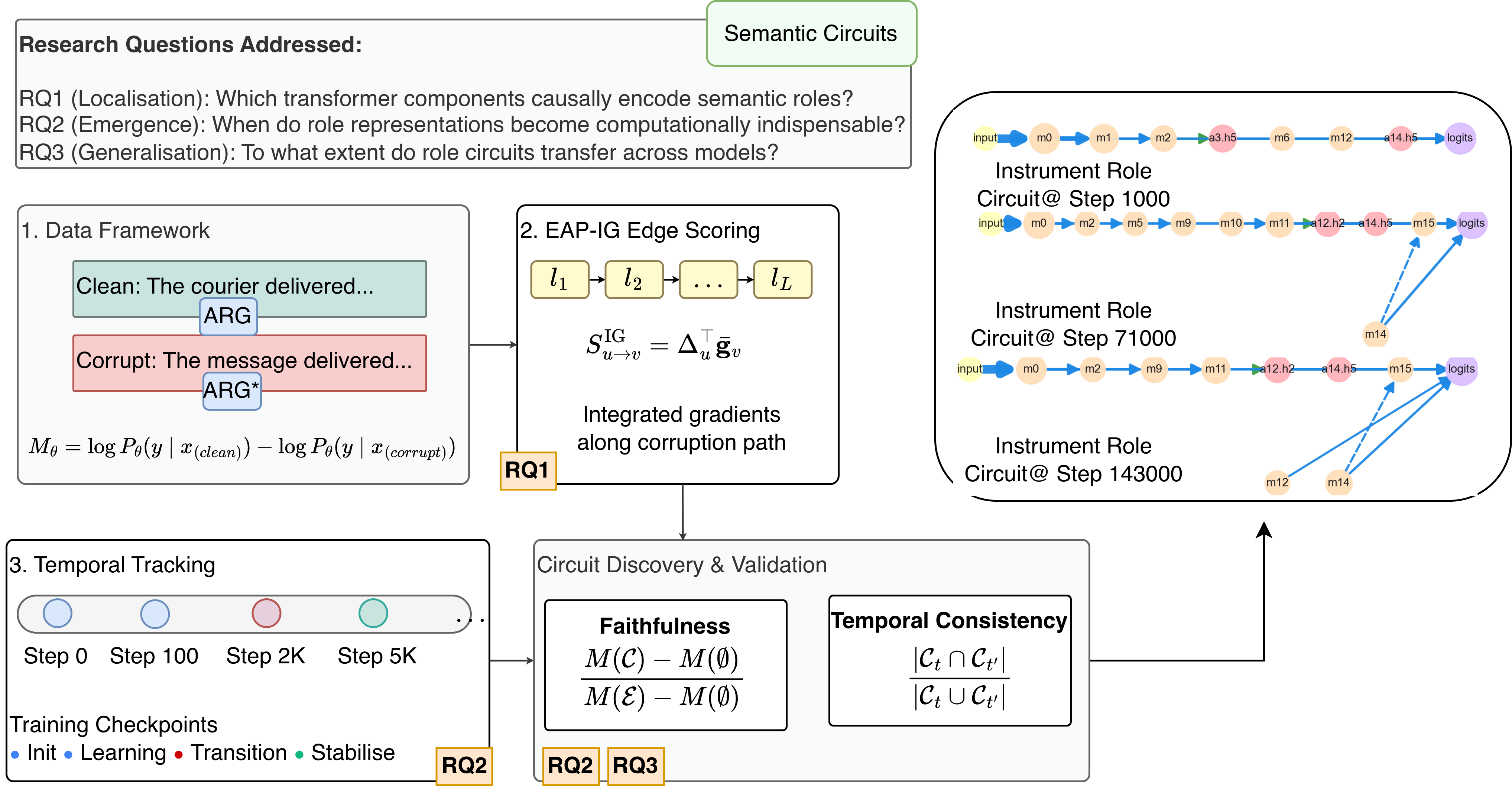}
    \caption{\textbf{COMPASS methodology.} It extracts and tracks the circuits that mediate semantic-role behaviour in LLMs, revealing where role-specific computation occurs and how it develops over training. (1) Role-cross minimal pairs isolate predicate–argument binding. (2) EAP-IG identifies edges whose interventions affect role predictions, producing sparse, causally functional subgraphs. (3) Temporal analysis follows these subgraphs across checkpoints to determine when their structure stabilises and when they become computationally indispensable.}    \label{fig:overview}
\end{figure*}
To address these questions, we introduce \textbf{COMPASS} (Compositional Predicate–Argument Semantic Structure; Fig.\ref{fig:overview}), a causal–temporal methodology that combines edge-attribution patching \citep{hanna2024have} with training-time circuit tracking\footnote{Supporting code and datasets are available at a public repository $<$ anonymised url $>$}. It investigates predicate–argument binding by isolating role-specific computation through contrastive prompts that differ only in their role-indicating scaffold (e.g., ``delivered the package \emph{to the} office'' vs. ``\emph{with the} truck''), what we term \emph{role-cross minimal pairs}, then tracks the causal pathways responsible for correct predictions. This approach disentangles the formation of circuit structure from its functional engagement and enables the evaluation of its transferability. We apply COMPASS across model scales and families to answer:

\begin{itemize}[leftmargin=*, itemsep=1pt, parsep=0pt, topsep=3pt]
    \item \textbf{RQ1 (Localisation):} Which model components causally encode semantic roles?
    \item \textbf{RQ2 (Emergence):} When do these circuits become computationally indispensable?
    \item \textbf{RQ3 (Generalisation):} To what extent do role circuits transfer across model scales and architectures?
\end{itemize}

We find that role-binding circuits localise to compact sets of attention heads and MLPs, with the results indicating that predicate–argument binding emerges through \emph{gradual refinement} rather than sudden reorganisation. Circuit presence does not guarantee immediate functional use: models appear to allocate semantic capacity early and exploit it only later. Larger models reuse similar components while wiring them differently. By demonstrating causally determined, functional circuits for shallow semantic structure, our method provides mechanistic evidence that LLMs acquire structured semantic representations and offers new avenues for targeted circuit editing and training-time intervention.

\paragraph{Contributions.} We summarise our contributions as: (i) introduce {COMPASS}, a causal–temporal method for discovering and tracking semantic-role circuits in LLMs; (ii) show that semantic-role information concentrates in small sets of components whose structural organisation stabilises before becoming functionally indispensable; and (iii) demonstrate partial cross-scale and cross-architecture transfer of these circuits.



\section{Related Work}\label{sec:related_work}

\paragraph{Mechanistic Interpretability (MI) and Circuit Discovery.} MI seeks to reverse-engineer neural network computations \citep{bereska2024mechanistic}. Foundational work analysed how attention heads, MLPs, and residual streams implement algorithms \citep{elhage2021mathematical}, leading to discoveries such as induction heads \citep{olsson2022context} and task-specific circuits \citep{wang2022interpretability, conmy2023towards}. Causal intervention methods form the core of modern circuit analysis: activation patching tests causal effects by swapping activations \citep{meng2022locating}, while gradient-based variants such as Attribution Patching (AtP) \citep{nanda2023attribution, syed-etal-2024-attribution} improve scalability but suffer from gradient saturation. $\text{AtP}^*$ \citep{kramar2024atp} introduces architectural fixes; Edge Attribution Patching (EAP) \citep{hanna2024have} attributes causal influence to individual edges, with EAP-IG mitigating saturation via Integrated Gradients. Parallel work on feature disentanglement uses Sparse Autoencoders \citep{templeton2024scaling, bricken2023monosemanticity} and transcoders \citep{dunefsky2024transcoders}, but they are surrogate-based approaches that introduce reconstruction errors and do not guarantee causal necessity \citep{gao2025scaling, kantamneni2025sparse}.

\paragraph{Training Dynamics and Compositional Emergence.}
Transformers often acquire capabilities through sharp transitions, including grokking-style shifts \citep{power2021grokking, nanda2023progress, aljaafari2025trace2} and phase changes in in-context learning \citep{wei2022emergent}. Recent work shows linguistic structure develops progressively: subspaces associated with syntax and semantics become more coherent over training \citep{muller-eberstein-etal-2023-subspace}, and task-specific circuits emerge in coordinated phases \citep{NEURIPS2024_47c7edad}. However, most analyses rely on probing or scalar metrics, leaving open whether transitions reflect emergence of causally functional circuits. To our knowledge, no prior work combines causal circuit localisation with systematic temporal tracking of compositional semantic representations across training and scale.

\paragraph{Semantic Understanding in Neural Language Models.}
While transformers achieve strong Semantic Role Labelling \citep{chen2025semanticrolelabelingsystematical}, most work examines accuracy rather than internal mechanisms. Probing studies suggest hierarchical linguistic knowledge organisation \citep{tenney2019bert, hewitt2019structural}, but probing reveals only linear separability, not causal involvement \citep{caucheteux2021disentangling, conia-navigli-2022-probing}. Circuit-level interpretability has mapped concrete behaviours such as IOI \citep{wang2022interpretability} but has not addressed whether transformers implement causally functional circuits for abstract predicate–argument relations, motivating our focus on semantic-role mechanisms.

\paragraph{Method Positioning.}
Probing scales efficiently but lacks causal grounding; intervention methods such as path patching \citep{GoldowskyDill2023LocalizingMB} and causal scrubbing \citep{chan2022causal} provide strong guarantees but are computationally prohibitive across many checkpoints. Surrogate-based approaches introduce reconstruction artefacts hindering temporal comparisons. We employ EAP-IG \citep{hanna2024have}, which provides path-specific causal attribution directly on the original model, extending it to a temporal setting to analyse when semantic-role circuits form, stabilise, and become functionally engaged across training and scale (detailed comparison of methods in App.~\ref{app:method-comparison}).
\section{Methodology}\label{sec:methodology}

\subsection{Semantic Role Circuits}\label{sec:task}

\paragraph{Computational graph representation.}
Following \citet{hanna2024have}, we view transformer computation as a directed acyclic graph $\mathcal{G}=(\mathcal{V},\mathcal{E})$ where nodes $u\in\mathcal{V}$ correspond to module outputs $u=(\text{type},\ell,h,i)$ with $\text{type}\in\{\text{AttnHead},\text{MLP}\}$, layer $\ell$, head $h$ (or $h=\varnothing$ for MLPs), and position $i$. Each $u$ outputs an activation $\mathbf{z}_u\in\mathbb{R}^{d_{\text{model}}}$ to the residual stream, and edges $(u\to v)\in\mathcal{E}$ denote residual connections.

\paragraph{Task: Role-conditioned continuation.}
To investigate predicate–argument binding, we adopt a role-conditioned continuation task grounded in frame semantics \citep{fillmore1976frame} and PropBank \citep{PropBank}. We vary the \emph{role-indicating scaffold} while keeping predicate and argument fillers fixed. If models represent abstract semantic roles, this behaviour should be mediated by \textbf{localised role circuits}~$\mathcal{C}^{(r)}$. We construct \textbf{role-cross minimal pairs} that differ only in the scaffold:
\begin{align*}
&\text{``The courier delivered the package \underline{to the}''} \\ 
&\to \text{\textsc{Goal}: ``office''} \\
&\text{``The courier delivered the package \underline{with the}''} \\
&\to \text{\textsc{Instrument}: ``truck''}
\end{align*}
Pairs use single-token fillers and enforce \textbf{token parity} ($|\text{toks}(x^{(r)})| = |\text{toks}(x^{(s)})|$). We retain only items where the model predicts correct continuations in both variants, ensuring circuits reflect functionally active behaviour (detailed in App.~\ref{app:data}).

\paragraph{Evaluation.} Task performance uses next-token accuracy:
\(\text{CNP}_{Acc} =\mathbb{E}_{(x^{(r)},y^{(r)})} \left[\mathds{1}\left[y^{(r)} = 
\argmax_v P_\theta(v \mid x^{(r)})\right]\right].\)
Circuit attribution uses the negative log-probability of the role-appropriate target: \(\mathcal{L}_{\text{CNP}}(x^{(r)}, y^{(r)}) = -\log P_\theta(y^{(r)} \mid x^{(r)}),\)
and EAP-IG attributes this loss to edges to identify components supporting role--appropriate predictions.

\subsection{COMPASS: Causal-Temporal Circuit Discovery}\label{sec:compass}
COMPASS integrates causal localisation with temporal tracking to recover role circuits $\mathcal{C}^{(r)}$ that exhibit (i) causal necessity, (ii) structural sparsity ($|\mathcal{C}^{(r)}| \ll |\mathcal{E}|$), and (iii) temporal stability. It identifies emergence times $(\hat{t}_c, t_{\text{cons}})$ characterising when circuits become functional. The procedure has three phases (full details of method and metrics are in App.~\ref{sec:metrics} and \ref{app:eap_ig}).

\paragraph{Phase 1: Causal localisation via EAP-IG.}
For each role pair $(x^{(r)},x^{(s)})$, EAP-IG \citep{hanna2024have} computes edge scores approximating their causal contribution to $\mathcal{L}_{\text{CNP}}$: \(S_{u\to v}^{\text{IG}} = \Delta_u^\top \bar{\mathbf{g}}_v, \quad \) where \(\Delta_u = \mathbf{z}_u^{(r)} - \mathbf{z}_u^{(s)},\)
and $\bar{\mathbf{g}}_v$ averages gradients along an IG interpolation path with $\alpha_k=k/m$ (we use $m{=}5$). We normalise by total mass: \(\tilde{S}_{u\to v}^{\text{IG}} = S_{u\to v}^{\text{IG}} / \sum_{e \in \mathcal{E}} |S_e^{\text{IG}}|\)
and extract the top-$K{=}200$ edges by $|\tilde{S}_{u\to v}^{\text{IG}}|$ at each checkpoint $t$ to define $\mathcal{C}_t^{(r)}$. Node importance is induced from incident edge mass: \(\text{Importance}^{(r)}_t(\ell,h) = \sum_{\substack{(u\to v) \in \mathcal{C}_t^{(r)} \\ v \in (\ell,h)}} |\tilde{S}_{u\to v}^{\text{IG}}|.\)

\paragraph{Phase 2: Temporal monitoring.}\label{sec:temporal}
At each checkpoint $t$, we compute causal and structural signals, including metrics from \citet{mueller2025mib}. \textit{Faithfulness} measures circuit indispensability: \(\text{Faithfulness}_t(\mathcal{C}) = {(M_t(\mathcal{C}) - M_t(\emptyset))}/{(M_t(\mathcal{E}) - M_t(\emptyset))},\) where $M_t(\mathcal{C})$ is $\text{CNP}_{acc}$ when only edges in $\mathcal{C}$ are active, $M_t(\mathcal{E})$ is full-model performance, and $M_t(\emptyset)$ is null baseline. \textit{Circuit persistence} is measured via Jaccard similarity: \(\text{Stability}_t(\mathcal{C}) = {|\mathcal{C}_t \cap \mathcal{C}_{t+\delta}|}/{|\mathcal{C}_t \cup \mathcal{C}_{t+\delta}|},\) along with Top-$K$ node mass and Gini coefficient.

\paragraph{Phase 3: Emergence detection.}
We define two temporal markers. \textit{Consolidation time} $t_{\text{cons}}$ marks when the top-$K$ node set stabilises: \(t_{\text{cons}} = \min\{t : \text{Stability}_t(\mathcal{C})\geq0.6 \text{ for } \geq 2 \text{ steps}\}.\)

\textit{Functional transition} $\hat{t}_c$ is the change-point in faithfulness, estimated by the best-fitting two-segment linear model via least-squares regression: \(\hat{t}_c = \argmax_{t} \, R^2\left(\text{PiecewiseLinear}(t)\right),
\)

with 95\% confidence intervals via bootstrap resampling ($n{=}1{,}000$; details in App.~\ref{app:eap_metrics}).

\subsection{Cross-Model Similarity}\label{sec:cross_model}
We assess circuit transferability across models by comparing structure and function via graph overlap and spectral geometry.

\paragraph{Structural overlap.}
Using the top-$K{=}30$ nodes and edges ranked by $|\tilde{S}|$, we compute Jaccard overlap between models $i,j$ for each role: \(J(V_i,V_j)={|V_i\cap V_j|}/{|V_i\cup V_j|}\), \(J(E_i,E_j)={|E_i\cap E_j|}/{|E_i\cup E_j|}.\)

\paragraph{Spectral similarity.}
To compare higher-order circuit geometry, we compute root-mean-square deviation between the smallest $k{=}16$ eigenvalues of the edge-weighted Laplacian using top-$K{=}50$ edges: \(d_{\text{spec}}(i,j)=\sqrt{\frac{1}{k}\sum_{m=1}^k(\lambda^{(i)}_m-\lambda^{(j)}_m)^2}.\)

Smaller $d_{\text{spec}}$ indicates similar information-flow geometry despite differing edge sets (full details in App.~\ref{app:eap_metrics}).
\section{Experimental Setup}\label{sec:experimental_setup}
\paragraph{Models.} We select models to satisfy three criteria: (i) dense training checkpoints to trace circuit formation, (ii) variation in scale to assess transferability, and (iii) architectural diversity to distinguish general strategies from implementation-specific artefacts. \textsc{Pythia} (14M, 410M, 1B) \citep{biderman2023pythia} provides comprehensive checkpoints, enabling fine-grained analysis of circuit emergence. To test whether these circuits persist beyond a single family, we include \textsc{LLaMA-1B} \citep{touvron2023llama}, which differs in tokenisation, training data, and architectural choices. This pairing separates training-time dynamics (within family) from architectural robustness (across families).

\paragraph{Datasets and Experimental Software.} \label{sec:exp_datasets}
We instantiate the role-conditioned continuation task (Sec.~\ref{sec:task}) across eight semantic roles spanning participant (\textsc{Beneficiary}, \textsc{Instrument}), directional (\textsc{Goal}, \textsc{Source}, \textsc{Path}), locative and temporal (\textsc{Location}, \textsc{Time}), and propositional (\textsc{Topic}) categories, providing broad coverage of predicate--argument structure types \citep{fillmore1976frame, palmer2005proposition, carnie2021syntax}. After filtering to retain only examples where models correctly predict role-appropriate targets in both contexts, we obtain approximately 6{,}000--8{,}000 pairs per model across roles (exact counts appear in Table~\ref{tab:dataset_summary}). Full dataset construction details, role categorisations, filtering procedures, and statistics are provided in App.~\ref{app:data}. Software and computation specifications appear in App.~\ref{app:software}.

\section{Results}\label{sec:results}
We address three research questions on semantic-role circuits across eight roles using \textsc{Pythia} (14M, 410M, 1B) and \textsc{LLaMA-1B}. For presentation clarity, we report main text results for four representative roles, \textsc{Beneficiary}, \textsc{Instrument}, \textsc{Location}, and \textsc{Time}; spanning participant, locative, and temporal categories. Results for the remaining roles (\textsc{Goal}, \textsc{Source}, \textsc{Path}, \textsc{Topic}) follow similar qualitative trends and are provided in App.~\ref{app:other-roles}. Unless otherwise specified, results refer to \textsc{Pythia-1B}. Comparisons across model scales and architectures are explicitly noted. Full results and metric definitions are given in App.~\ref{app:full-results} and~\ref{sec:metrics}.


\subsection{Localisation of Role Circuits (RQ1)}\label{sec:rq1}
We find highly localised circuits for all tested roles, with the top 20 nodes capturing 89--92\% of attribution mass at convergence (Tab.~\ref{tab:rq2-sparsity}). Despite this shared sparsity, roles exhibit qualitatively distinct architectures and developmental trajectories.

\begin{figure}[h!]
    \centering
    \includegraphics[width=\linewidth]{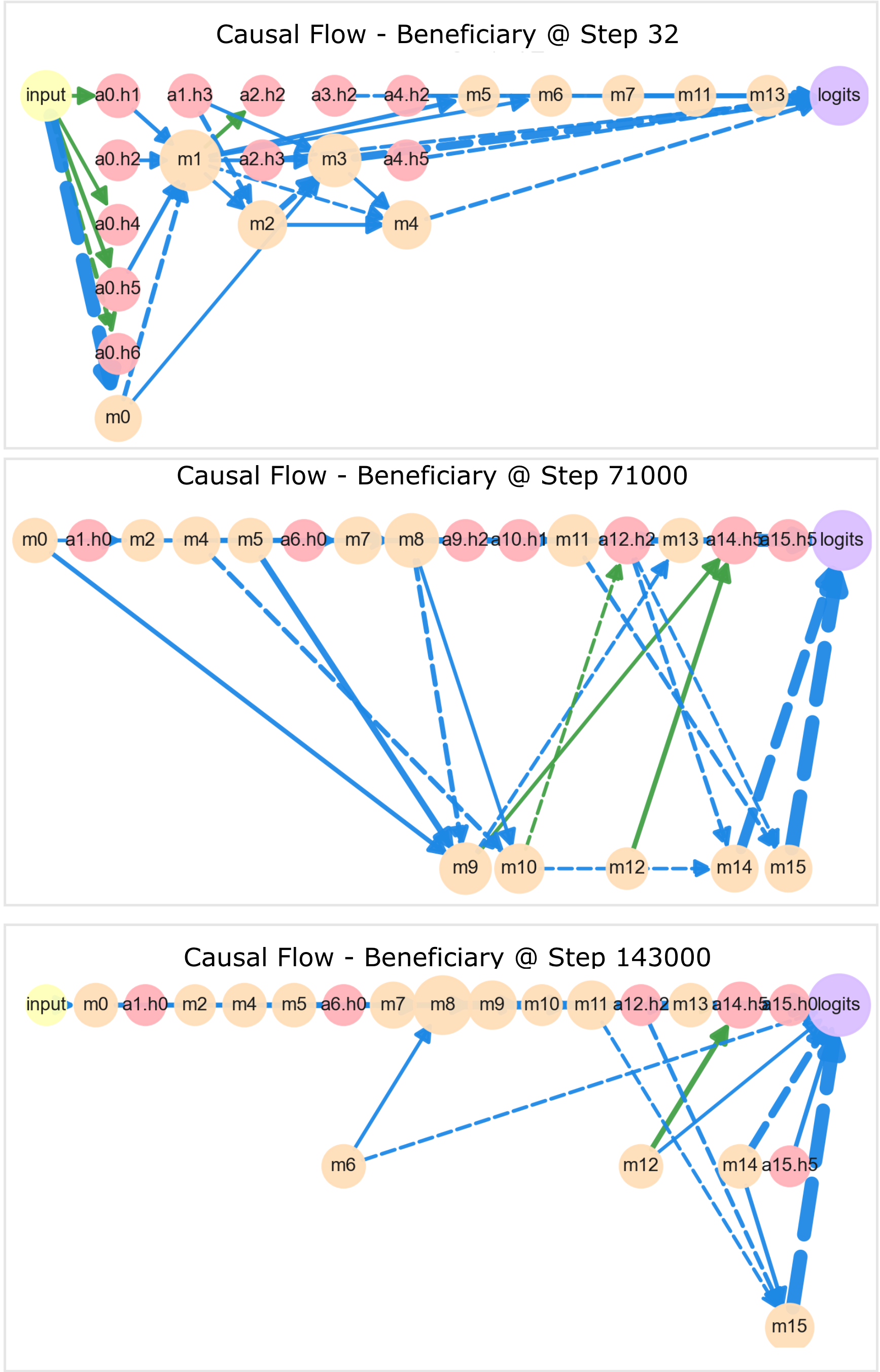}
    \caption{\textbf{Evolution of the \textsc{Beneficiary} circuit across training.} The circuit reorganises from early exploration (step 32), through mid-training feature extraction (step 71K), to its final architecture (step 143K). Edge colour encodes operation type ({\color{NavyBlue} blue}: residual; {\color{Green} green}: value composition; dashed: negative), and edge width indicates attribution magnitude.}
\label{fig:causal-flow-grid}
\end{figure}

\begin{figure*}
    \centering
    \includegraphics[width=\linewidth]{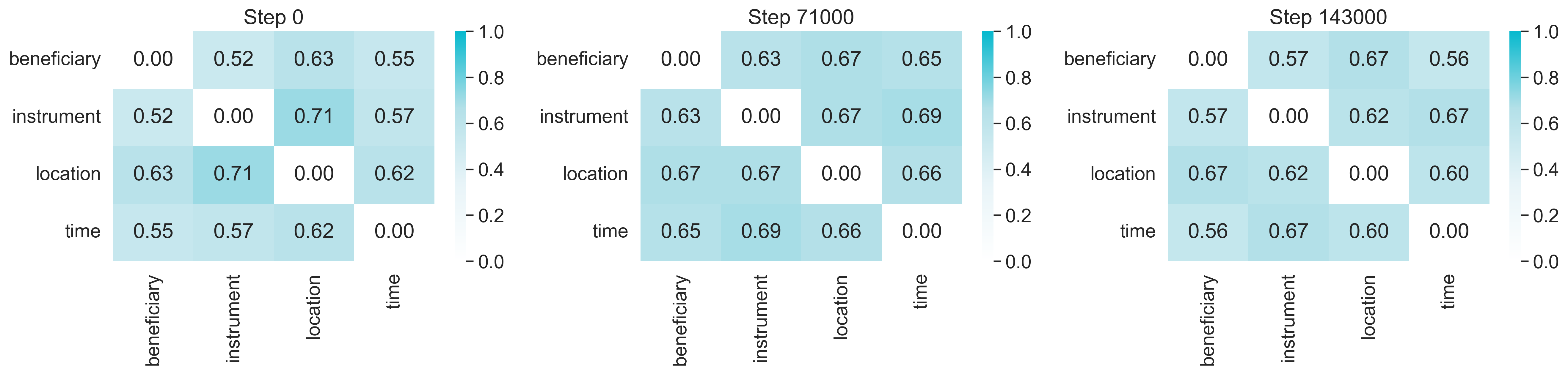}
    \caption{\textbf{Cross-role overlap of high-importance components over training (\textsc{Pythia–1B}).} Overlap remains consistently low across training stages, indicating roles recruit largely distinct component sets, supporting circuit differentiation rather than shared mechanisms.}
    \label{fig:role-overlap}
\end{figure*}

\paragraph{Circuit architectures vary by role.}
Causal-flow analysis (Fig.~\ref{fig:causal-flow-grid}) reveals systematic cross-role differences. \textsc{Beneficiary} develops the most complex architecture, combining early attention, extensive MLP branching, and rich late-layer connectivity with multiple value-composition operations. In contrast, \textsc{Time} shifts from early-layer attention to convergent mid-to-late processing; \textsc{Instrument} stabilises to a balanced hybrid circuit; and \textsc{Location} transitions from mid-training expansion to more distributed integration. Low to mid Jaccard overlap across roles (Fig.~\ref{fig:role-overlap}) confirms these correspond to distinct, role-specific subgraphs. Full circuit evolution for additional roles in App.~\ref{sec:circuit_heterogeneity}.

\begin{table}[ht]
\centering
\resizebox{\columnwidth}{!}{
\begin{tabular}{lccccc}
\toprule
Role & T-5 & T-10 & T-20 & $k$ for 80/90/95\%  & Gini (mass)\\
\midrule
\textsc{Beneficiary} & 0.395 &	0.611 &	\textbf{0.906} & 16 / 20 / 25  & \textbf{0.439} \\
\textsc{Instrument}  & 0.492& 0.688& \textbf{0.917} & 14 / 19 / 23 & \textbf{0.489} \\
\textsc{Location}    & 0.483&0.673 & \textbf{0.897} & 15 / 21 / 25 & \textbf{0.519} \\
\textsc{Time}        & 0.466& 0.664 & \textbf{0.910} & 15 / 20 / 24 & \textbf{0.475}\\
\bottomrule
\end{tabular}
}
\caption{\textbf{Final-step concentration profiles of role circuits (step 143K).} For each role, we report cumulative attribution mass captured by the top (T-$k$) nodes (T-5, T-10, T-20), the smallest $k$ achieving 80/90/95\% total mass, and the Gini coefficient over attribution mass. Together, these metrics show that role circuits are highly sparse, with a small set of nodes dominating computation, while differences in Gini reflect variation in how evenly mass is distributed beyond the top contributors.}
\label{tab:rq2-sparsity}
\end{table}

\begin{figure*}[ht]
    \centering
    \includegraphics[width=0.95\linewidth]{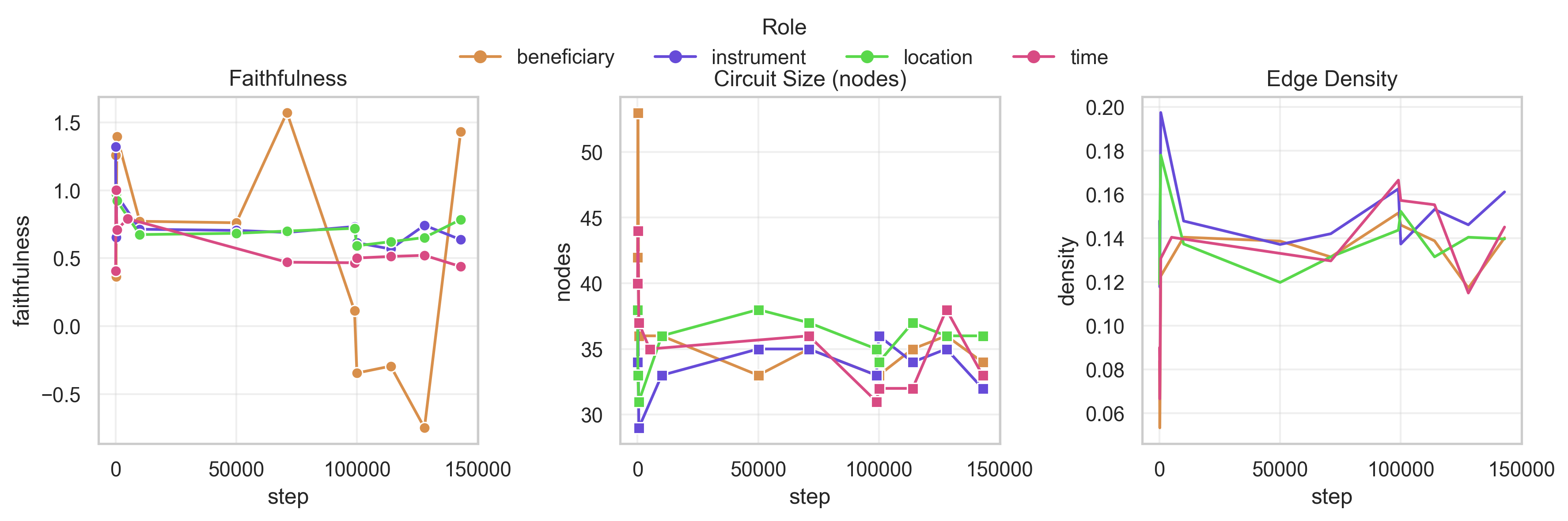}
    \caption{\textbf{Structural and functional dynamics of role circuits across training.} Faithfulness (left) shows pronounced role-dependent volatility. In contrast, structural metrics evolve smoothly: circuit size (middle) contracts gradually, and edge density (right) rises or stabilises, showing structure consolidates early and steadily while functional engagement remains variable.}
    \label{fig:role-evolution}
\end{figure*}

\paragraph{Roles exhibit distinct developmental trajectories.}
Temporal analysis (Fig.~\ref{fig:role-evolution}) shows heterogeneous emergence dynamics. \textsc{Instrument} starts with high faithfulness, undergoes a sharp decline, then stabilises with edge density dipping before rising, suggesting late consolidation into denser pathways. \textsc{Beneficiary} exhibits the most pronounced volatility: faithfulness spikes sharply at step 50k, drops sharply, partially recovers; circuit size decreases early but re-expands in the final 70K steps, yielding a distinctive ``expand--contract'' pattern. \textsc{Location} and \textsc{Time} maintain relatively stable faithfulness after initial drops, with edge density showing moderate fluctuations, consistent with gradual strengthening of connectivity. Circuit sizes contract from initialisation (29--54 nodes) to convergence (31--36 nodes) across all roles, with \textsc{Beneficiary} showing the largest initial size and most rapid contraction.

\paragraph{Key findings.}
\textbf{(i) Highly concentrated circuits with role-specific structure.} All roles converge to compact circuits (89--92\% mass in $\leq$20 nodes) with role-specific architectures confirmed by low cross-role overlap. \textbf{(ii) Continuous refinement, not discrete transitions.} Structural metrics (circuit size, edge density) evolve smoothly throughout training, indicating continuous refinement rather than discrete transitions. \textbf{(iii) Structure--function dissociation.} Faithfulness exhibits pronounced non-monotonicity even as circuits structurally consolidate. This demonstrates that circuits can be structurally well-formed yet functionally but temporarily underutilised. \textbf{(iv) Role-specific stabilisation timelines.} Roles show distinct functional trajectories: \textsc{Beneficiary} exhibits extreme volatility, \textsc{Instrument} sharp early decline then stability, \textsc{Location} and \textsc{Time} moderate stability after initial adjustment.

\begin{table}[ht]
\centering
\small
\begin{tabular}{lcc}
\toprule
{Role} & $t_{\mathrm{ind}}$ (steps) & $t_{\mathrm{cons}}$ (steps) \\
\midrule
\textsc{Beneficiary} & --  &  50{,}000  \\
\textsc{Instrument}  & 32 & 128 \\
\textsc{Location}    & 128  & 128  \\
\textsc{Time}        & 512 & 5{,}000  \\
\bottomrule
\end{tabular}
\caption{\textbf{Emergence timings for role circuits.} $t_{\mathrm{ind}}$ marks earliest step where ablation consistently harms performance; $t_{\mathrm{cons}}$ marks when structure stabilises. All roles consolidate within $\sim$2k steps, but indispensability varies widely.}
\label{tab:rq2-tables}
\end{table}

\subsection{Emergence Dynamics (RQ2)}\label{sec:rq2}
We find that role-binding circuits emerge via \emph{continuous refinement} rather than discrete phase transitions. Roles become causally indispensable early (0--512 steps), but structural consolidation unfolds over tens of thousands of steps, producing pronounced \emph{structure--function dissociation}.

\paragraph{Early indispensability with heterogeneous emergence.} Roles become indispensable at different stages (Tab.~\ref{tab:rq2-tables}): \textsc{Instrument} by step 32, \textsc{Location} by step 128, and \textsc{Time} by step 512. \textsc{Beneficiary} exhibits fluctuating improvement in faithfulness and never exceeds our conservative indispensability threshold ($\mu + \sigma$), leaving $t_{\text{ind}}$ undefined and indicating usefulness without provable necessity under this criterion. This dissociation likely reflects differences in cue salience, frequency, and computational demands. Architectural heterogeneity across roles is detailed in App.~\ref{sec:circuit_heterogeneity}.

\paragraph{Continuous sparsification without phase transitions.}
Piecewise linear regression on Top-$K$ mass yields extremely wide change-point confidence intervals (CIs; Tab.~\ref{tab:rq2-changepoint-topk}), ruling out discrete transitions. This observation contrasts with grokking, where change-points are sharp, localised and abrupt \citep{power2021grokking, nanda2023progress}. Instead, structural metrics mostly evolve smoothly throughout training (Fig.~\ref{fig:role-evolution}), consistent with gradual sparsification, with some per role specification like \textsc{Location} exhibiting mild ``expand--contract'' patterns. Final concentration is extreme: Top-20 nodes capture 89.7--91.7\% of mass, with 95\% coverage in just 23--25 nodes (Tab.~\ref{tab:rq2-sparsity}).

\begin{table}[t]
\centering
\small
\begin{tabular}{lcc}
\toprule
{Role} & $\hat{t}_c$ (steps) & 95\% CI \\
\midrule
\textsc{Beneficiary}& 10{,}000   & [128, 10{,}000] \\
\textsc{Instrument} & 50{,}000 & [128, 10{,}000] \\
\textsc{Location}   & 10{,}000   & [128, 10{,}000] \\
\textsc{Time}       & 5{,}000   & [128, 114{,}000] \\
\bottomrule
\end{tabular}
\caption{\textbf{Change-point estimates for Top-$K$ node mass} via piecewise linear regression. Extremely wide confidence intervals indicate gradual, continuous evolution without discrete phase transitions.}
\label{tab:rq2-changepoint-topk}
\end{table}

\paragraph{Structure--function dissociation.} Although structural metrics evolve smoothly, faithfulness trajectories exhibit pronounced non-monotonicity (Fig.~\ref{fig:role-evolution}, left). Structural consolidation precedes functional indispensability by several steps (Tab.~\ref{tab:rq2-tables}: $t_{\text{cons}}{=}128$ vs.\ $t_{\text{ind}}{=}32$--512), demonstrating that \emph{circuit presence does not guarantee circuit engagement}. 

\paragraph{Robustness to scaffold variation and frequency effects.}
To test whether circuits reflect abstract role binding rather than memorised lexical scaffolds, we evaluate within-role paraphrases for \textsc{Location} and \textsc{Instrument}. Faithfulness and sparsity remain largely unchanged under paraphrase (App.~\ref{app:paraphrase-controls}). Moreover, circuit properties at convergence show weak correlations with (i) per-role filtered sample size and (ii) scaffold frequency in the pretraining corpus (App.~\ref{app:role_frequency_analysis}), suggesting the observed architectural differences are not driven by dataset volume or training-signal strength alone.

\paragraph{Key findings.}
\textbf{(i) Continuous refinement throughout training.} Structural metrics evolve smoothly, with wide change-point CIs indicating no discrete phase shifts. \textbf{(ii) Early indispensability, prolonged consolidation.} Roles become indispensable within 0--512 steps, while structural refinement continues for over 143K steps as attribution mass redistributes. \textbf{(iii) Structure--function dissociation and delayed engagement.} Circuits stabilise structurally many steps before becoming functionally indispensable; faithfulness shows crashes and recoveries even as sparsity increases, indicating that \emph{circuit presence does not mean engagement}. \textbf{(iv) Role-specific developmental trajectories.} Despite shared qualitative patterns, roles exhibit distinct functional dynamics reflecting role-specific computational demands.

\subsection{Cross-Scale and Cross-Family Generalisation (RQ3)} \label{sec:rq3}
We find moderate structural conservation (24–51\% node overlap) across models. They converge on shared \emph{functional vocabularies} while implementing divergent \emph{routing patterns}: component reuse exceeds connection reuse by $\sim$2:1. Spectral analysis reveals functional alignment despite topological divergence, with small eigenvalue distances ($<0.02$) coexisting with 76–88\% edge mismatch.

\paragraph{Cross-scale correspondence.}
Node-level overlap increases with scale proximity (Tab.~\ref{tab:cross-scale-summary}): 14M$\leftrightarrow$410M yields $J_V{=}0.24$, 14M$\leftrightarrow$1B reaches $J_V{=}0.29$, and 410M$\leftrightarrow$1B achieves $J_V{=}0.42$, the strongest within-family match. Edge-level overlap remains low ($J_E{\approx}0.11$--0.17), indicating models reuse similar high-importance nodes but wire them differently. Spectral distances decrease with scale ($d_{\text{spec}}{=}0.12{\rightarrow}0.10{\rightarrow}0.01$), suggesting progressive geometric refinement: larger models realise similar information-flow patterns with increasingly aligned connectivity, reflecting shared training conditions (e.g., corpus, optimiser) and architectural continuity within families.

\begin{table}[t]
\centering
\footnotesize
\setlength{\tabcolsep}{2pt}
\begin{tabular*}{\columnwidth}{@{\extracolsep{\fill}}lccc@{}}
\toprule
Model pair & Node J. & Edge J. & $d_{\text{spec}}$ \\
\midrule
\textsc{Pythia–14M} $\leftrightarrow$ \textsc{Pythia–410M}  & 0.24 & 0.11 & 0.12 \\
\textsc{Pythia–14M} $\leftrightarrow$ \textsc{Pythia–1B}   & 0.29 & 0.12 & 0.10 \\
\textsc{Pythia–410M} $\leftrightarrow$ \textsc{Pythia–1B}  & 0.42 & 0.17 & 0.01 \\
\textsc{Pythia–1B} $\leftrightarrow$ \textsc{LLaMA–1B}     & 0.51 & 0.14 & 0.02 \\
\bottomrule
\end{tabular*}
\caption{\textbf{Cross-scale and -family similarity of role circuits.} Node-level overlap increases with model scale while edge-level overlap remains low. Cross-family comparison shows the highest node overlap. Values report median Top-$K$ Jaccard ($K{=}30$) and spectral distance. Lower $d_{\text{spec}}$ indicates higher functional similarity.}
\label{tab:cross-scale-summary}
\end{table}

\paragraph{Cross-family correspondence.}
The \textsc{Pythia-1B}$\leftrightarrow$\textsc{LLaMA-1B} comparison exhibits the \emph{highest node overlap} ($J_V{=}0.51$), exceeding even the closest within-Pythia pair (410M$\leftrightarrow$1B: $J_V{=}0.42$). This suggests architectural constraints at the 1B scale bias both models toward similar component sets despite different pretraining corpora and architectural choices. However, edge-level overlap remains modest ($J_E{=}0.14$) and spectral distance ($d_{\text{spec}}{=}0.02$) slightly lower than the best within-family match. This ``shared components, divergent wiring'' pattern indicates models converge on a common functional vocabulary while implementing distinct routing schemas shaped by architectural and training differences.

\begin{figure}[ht]
    \centering
    \includegraphics[width=\linewidth]{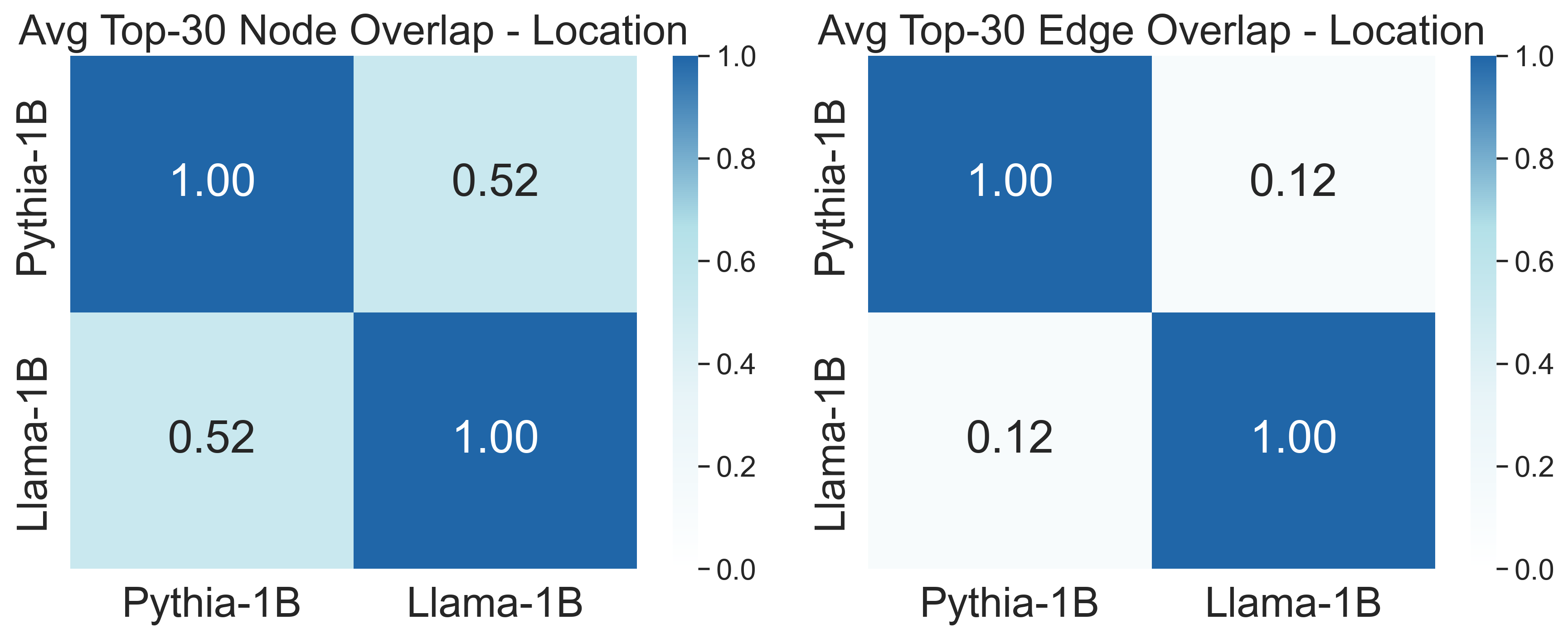}
    \includegraphics[width=\linewidth]{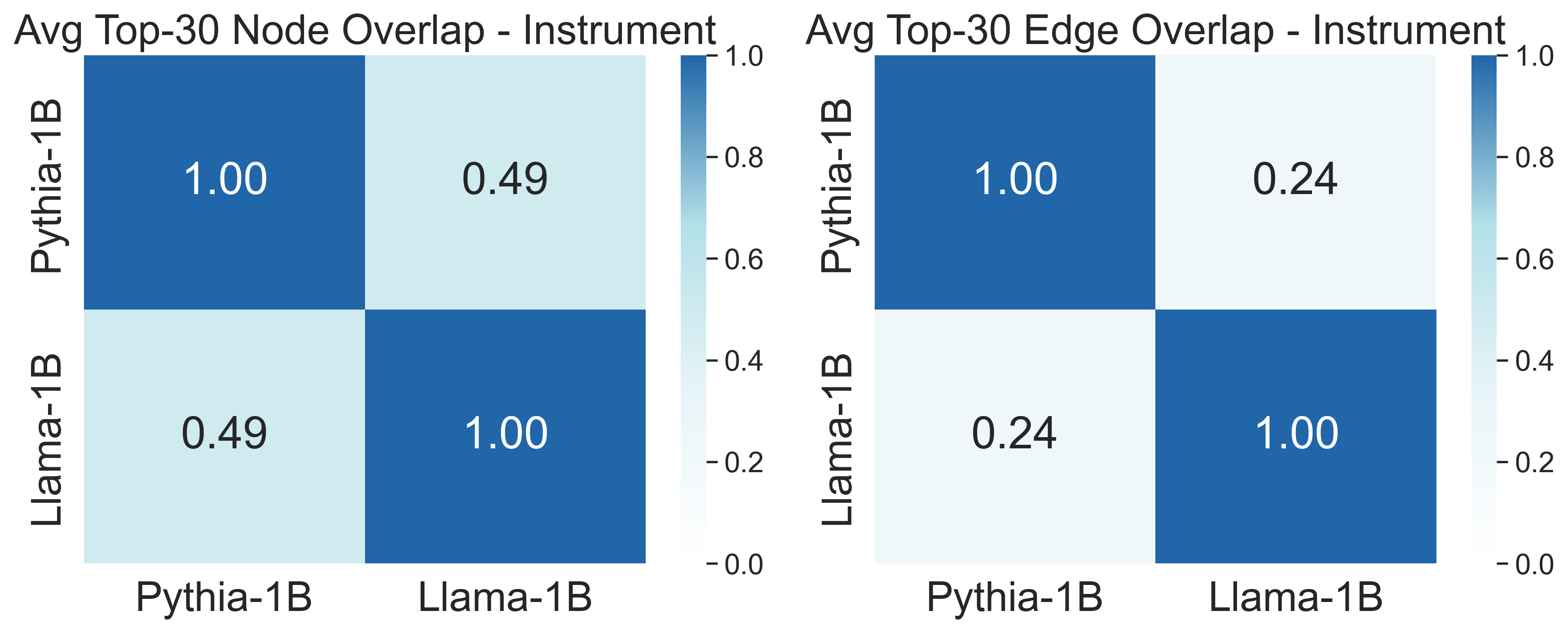}
    \caption{\textbf{Cross-family correspondence for Location (top) and Instrument (bottom).} Node sets align substantially more than edges between \textsc{Pythia–1B} and \textsc{LLaMA–1B}, suggesting shared component selection but model-specific routing.}
    \label{fig:cross-scale-heatmaps}
\end{figure}

\paragraph{Role-specific patterns.}
Per-role analysis (Fig.~\ref{fig:cross-scale-heatmaps}) reveals heterogeneous cross-family transferability. \textsc{Location} shows higher node overlap (52\%) than \textsc{Instrument} (49\%) between Pythia-1B and LLaMA-1B, but lower edge overlap (12\% vs 24\%), suggesting \textsc{Location}'s distributed architecture reuses components while reorganising connections more substantially. \textsc{Instrument} maintains stronger edge preservation (24\%), consistent with more stable routing patterns. Across both roles, node-level overlap consistently exceeds edge-level overlap by $\sim$2–4$\times$, reinforcing that component reuse dominates connection reuse even across architectural families. Cross-family overlap often exceeds these cross-scale values within the Pythia family for nearby scale pairs, with full per-role results and scale-specific patterns in App.~\ref{app:full-results}.

\paragraph{Key findings.} \textbf{(i) Moderate structural conservation across scales and families.} High-importance nodes show 24–51\% overlap; cross-family correspondence (51\%) exceeds strongest within-family pairing (42\%), indicating circuits reflect both task demands and model-specific factors. \textbf{(ii) Component reuse exceeds connection reuse.} Models converge on shared functional vocabularies (which components matter) while diverging in routing structure. \textbf{(iii) Spectral alignment despite topological divergence.} Small spectral distances coexist with large edge differences, showing models realise similar information--flow geometry via distinct connectivity. \textbf{(iv) Scale-dependent refinement within families.} Spectral distance decreases monotonically across \textsc{Pythia} scales; cross-family increase reflects architectural differences.
\section{Conclusion}\label{sec:discussion}
Our results characterise semantic-role circuits along three dimensions. \emph{RQ1} shows that role information localises into compact, role-specific subgraphs. \emph{RQ2} indicates gradual emergence: functional importance can appear early while structural properties evolve, with role-dependent faithfulness. \emph{RQ3} reveals partial cross-scale conservation, with components reused across models but connected through different routing schemes. {Mechanistically}, the absence of sharp transitions suggests that circuits are shaped by sustained optimisation rather than being ``switched on'' once competence appears. From a linguistic standpoint, these circuits approximate role–filler bindings, central to predicate–argument semantics (\emph{who did what to whom, when, with what}), implying partially modular causal mechanisms rather than purely diffuse heuristics. {COMPASS} extends MI temporally (when circuits appear and engage), cross-scale (how they persist across sizes/families), and semantically (predicate--argument relations beyond lexical cues). The presence of compact, causally functional circuits and their partial transfer across models suggests these mechanisms reflect underlying task structure rather than mere memorising surface co-occurrences. For safety and alignment, localising such mechanisms may enable targeted interventions on specific subgraphs, complementing training-time approaches that promote compositionality \citep{aljaafari-etal-2025-carma}. Heterogeneous emergence timelines further motivate investigating curricula or early-stopping strategies that prioritise stabilising particular semantic capabilities.

\section*{Limitations} This study focuses on English and a subset of roles; whether similar circuits arise in typologically diverse languages or richer role inventories remains unknown. Wide change-point confidence intervals reflect smooth structural trajectories, but more sensitive emergence metrics may yield finer resolution. We analyse decoder-only architectures; encoder–decoder models may exhibit different patterns. Future work should extend to multilingual and multimodal models, study interactions between semantic-role circuits and syntax or coreference mechanisms, and test whether analogous patterns appear for other semantic abstractions such as quantification, modality, or negation.

\section*{Ethical considerations}
This work aims to improve the interpretability of semantic processing in LLMs through circuit discovery methods. While understanding internal mechanisms benefits safety research and model development, we acknowledge potential risks in localising computational pathways that could be exploited for targeted interventions. Our analysis focuses on controlled semantic tasks with comprehensive validation across training and scales to ensure responsible development and transparent reporting of findings.
\bibliography{custom}

@inproceedings{
hanna2024have,
title={Have Faith in Faithfulness: Going Beyond Circuit Overlap When Finding Model Mechanisms},
author={Michael Hanna and Sandro Pezzelle and Yonatan Belinkov},
booktitle={ICML 2024 Workshop on Mechanistic Interpretability},
year={2024},
url={https://openreview.net/forum?id=grXgesr5dT}
}

@inproceedings{power2021grokking,
  title={Grokking: Generalization beyond overfitting on small algorithmic datasets},
  author={Power, Alethea and Burda, Yuri and Edwards, Harri and Babuschkin, Igor and Misra, Vedant},
  booktitle={ICLR MATH-AI Workshop},
  year={2021}
}

@article{meng2022locating,
  title={Locating and editing factual associations in GPT},
  author={Meng, Kevin and Bau, David and Andonian, Alex and Belinkov, Yonatan},
  journal={Advances in neural information processing systems},
  volume={35},
  pages={17359--17372},
  year={2022}
}

@article{elhage2021mathematical,
   title={A Mathematical Framework for Transformer Circuits},
   author={Elhage, Nelson and Nanda, Neel and Olsson, Catherine and Henighan, Tom and Joseph, Nicholas and Mann, Ben and Askell, Amanda and Bai, Yuntao and Chen, Anna and Conerly, Tom and DasSarma, Nova and Drain, Dawn and Ganguli, Deep and Hatfield-Dodds, Zac and Hernandez, Danny and Jones, Andy and Kernion, Jackson and Lovitt, Liane and Ndousse, Kamal and Amodei, Dario and Brown, Tom and Clark, Jack and Kaplan, Jared and McCandlish, Sam and Olah, Chris},
   year={2021},
   journal={Transformer Circuits Thread},
   note={https://transformer-circuits.pub/2021/framework/index.html}
}

@inproceedings{
    wang2022interpretability,
    title={Interpretability in the Wild: a Circuit for Indirect Object Identification in {GPT}-2 Small},
    author={Kevin Ro Wang and Alexandre Variengien and Arthur Conmy and Buck Shlegeris and Jacob Steinhardt},
    booktitle={The Eleventh International Conference on Learning Representations },
    year={2023},
    url={https://openreview.net/forum?id=NpsVSN6o4ul}
}

@article{olsson2022context,
  title={In-context learning and induction heads},
  author={Olsson, Catherine and Elhage, Nelson and Nanda, Neel and Joseph, Nicholas and DasSarma, Nova and Henighan, Tom and Mann, Ben and Askell, Amanda and Bai, Yuntao and Chen, Anna and others},
  journal={arXiv preprint arXiv:2209.11895},
  year={2022}
}

@inproceedings{
nanda2023progress,
title={Progress measures for grokking via mechanistic interpretability},
author={Neel Nanda and Lawrence Chan and Tom Lieberum and Jess Smith and Jacob Steinhardt},
booktitle={The Eleventh International Conference on Learning Representations },
year={2023},
url={https://openreview.net/forum?id=9XFSbDPmdW}
}

@article{
wei2022emergent,
title={Emergent Abilities of Large Language Models},
author={Jason Wei and Yi Tay and Rishi Bommasani and Colin Raffel and Barret Zoph and Sebastian Borgeaud and Dani Yogatama and Maarten Bosma and Denny Zhou and Donald Metzler and Ed H. Chi and Tatsunori Hashimoto and Oriol Vinyals and Percy Liang and Jeff Dean and William Fedus},
journal={Transactions on Machine Learning Research},
issn={2835-8856},
year={2022},
url={https://openreview.net/forum?id=yzkSU5zdwD},
note={Survey Certification}
}

@inproceedings{biderman2023pythia,
  title={Pythia: A suite for analyzing large language models across training and scaling},
  author={Biderman, Stella and Schoelkopf, Hailey and Anthony, Quentin Gregory and Bradley, Herbie and O’Brien, Kyle and Hallahan, Eric and Khan, Mohammad Aflah and Purohit, Shivanshu and Prashanth, USVSN Sai and Raff, Edward and others},
  booktitle={International Conference on Machine Learning},
  pages={2397--2430},
  year={2023},
  organization={PMLR}
}

@article{aljaafari2025trace2,
  title={TRACE for Tracking the Emergence of Semantic Representations in Transformers},
  author={Aljaafari, Nura and Carvalho, Danilo S and Freitas, Andr{\'e}},
  journal={arXiv preprint arXiv:2505.17998},
  year={2025}
}

@article{
bereska2024mechanistic,
title={Mechanistic Interpretability for {AI} Safety - A Review},
author={Leonard Bereska and Stratis Gavves},
journal={Transactions on Machine Learning Research},
issn={2835-8856},
year={2024},
month   = {Aug},
url={https://openreview.net/forum?id=ePUVetPKu6},
}

@article{conmy2023towards,
  title={Towards automated circuit discovery for mechanistic interpretability},
  author={Conmy, Arthur and Mavor-Parker, Augustine and Lynch, Aengus and Heimersheim, Stefan and Garriga-Alonso, Adri{\`a}},
  journal={Advances in Neural Information Processing Systems},
  volume={36},
  pages={16318--16352},
  year={2023}
}

@misc{nanda2023attribution,
  author       = {Nanda, Neel},
  title        = {Attribution Patching: Activation Patching at Industrial Scale},
  year         = {2023},
  howpublished = {\url{https://www.neelnanda.io/mechanistic-interpretability/attribution-patching}},
}

@inproceedings{syed-etal-2024-attribution,
    title = "Attribution Patching Outperforms Automated Circuit Discovery",
    author = "Syed, Aaquib  and
      Rager, Can  and
      Conmy, Arthur",
    editor = "Belinkov, Yonatan  and
      Kim, Najoung  and
      Jumelet, Jaap  and
      Mohebbi, Hosein  and
      Mueller, Aaron  and
      Chen, Hanjie",
    booktitle = "Proceedings of the 7th BlackboxNLP Workshop: Analyzing and Interpreting Neural Networks for NLP",
    month = nov,
    year = "2024",
    address = "Miami, Florida, US",
    publisher = "Association for Computational Linguistics",
    url = "https://aclanthology.org/2024.blackboxnlp-1.25/",
    doi = "10.18653/v1/2024.blackboxnlp-1.25",
    pages = "407--416"
}

@article{palmer2005proposition,
author = {Palmer, Martha and Gildea, Daniel and Kingsbury, Paul},
title = {The Proposition Bank: An Annotated Corpus of Semantic Roles},
year = {2005},
issue_date = {March 2005},
publisher = {MIT Press},
address = {Cambridge, MA, USA},
volume = {31},
number = {1},
issn = {0891-2017},
url = {https://doi.org/10.1162/0891201053630264},
doi = {10.1162/0891201053630264},
journal = {Comput. Linguist.},
month = mar,
pages = {71–106},
numpages = {36}
}

@misc{chen2025semanticrolelabelingsystematical,
      title={Semantic Role Labeling: A Systematical Survey}, 
      author={Huiyao Chen and Meishan Zhang and Jing Li and Min Zhang and Lilja Øvrelid and Jan Hajič and Hao Fei},
      year={2025},
      eprint={2502.08660},
      archivePrefix={arXiv},
      primaryClass={cs.CL},
      url={https://arxiv.org/abs/2502.08660}, 
}

@InProceedings{cheng2024potentiallimitationsllmscapturing,
author="Cheng, Ning
and Yan, Zhaohui
and Wang, Ziming
and Li, Zhijie
and Yu, Jiaming
and Zheng, Zilong
and Tu, Kewei
and Xu, Jinan
and Han, Wenjuan",
editor="Huang, De-Shuang
and Zhang, Xiankun
and Zhang, Qinhu",
title="Potential and Limitations of LLMs in Capturing Structured Semantics: A Case Study on SRL",
booktitle="Advanced Intelligent Computing Technology and Applications",
year="2024",
publisher="Springer Nature Singapore",
address="Singapore",
pages="50--61",
isbn="978-981-97-5663-6"
}

@inproceedings{conia-navigli-2022-probing,
    title = "Probing for Predicate Argument Structures in Pretrained Language Models",
    author = "Conia, Simone  and
      Navigli, Roberto",
    editor = "Muresan, Smaranda  and
      Nakov, Preslav  and
      Villavicencio, Aline",
    booktitle = "Proceedings of the 60th Annual Meeting of the Association for Computational Linguistics (Volume 1: Long Papers)",
    month = may,
    year = "2022",
    address = "Dublin, Ireland",
    publisher = "Association for Computational Linguistics",
    url = "https://aclanthology.org/2022.acl-long.316/",
    doi = "10.18653/v1/2022.acl-long.316",
    pages = "4622--4632"
}

@article{chan2022causal, 
	title={Causal scrubbing, a method for rigorously testing interpretability hypotheses},
	author={Chan, Lawrence and Garriga-Alonso, Adrià and Goldwosky-Dill, Nicholas and Greenblatt, Ryan and Nitishinskaya, Jenny and Radhakrishnan, Ansh and Shlegeris, Buck and Thomas, Nate},
	year={2022},
	journal={AI Alignment Forum},
	url={https://www.alignmentforum.org/posts/JvZhhzycHu2Yd57RN/causal-scrubbing-a-method-for-rigorously-testing}
}

@article{GoldowskyDill2023LocalizingMB,
  title={Localizing Model Behavior with Path Patching},
  author={Nicholas Goldowsky-Dill and Chris MacLeod and Lucas Jun Koba Sato and Aryaman Arora},
  journal={ArXiv},
  year={2023},
  volume={abs/2304.05969},
  url={https://api.semanticscholar.org/CorpusID:258079237}
}

@article{bricken2023monosemanticity,
       title={Towards Monosemanticity: Decomposing Language Models With Dictionary Learning},
       author={Bricken, Trenton and Templeton, Adly and Batson, Joshua and Chen, Brian and Jermyn, Adam and Conerly, Tom and Turner, Nick and Anil, Cem and Denison, Carson and Askell, Amanda and Lasenby, Robert and Wu, Yifan and Kravec, Shauna and Schiefer, Nicholas and Maxwell, Tim and Joseph, Nicholas and Hatfield-Dodds, Zac and Tamkin, Alex and Nguyen, Karina and McLean, Brayden and Burke, Josiah E and Hume, Tristan and Carter, Shan and Henighan, Tom and Olah, Christopher},
       year={2023},
       journal={Transformer Circuits Thread},
       url={https://transformer-circuits.pub/2023/monosemantic-features/index.html}
    }

@article{templeton2024scaling,
       title={Scaling Monosemanticity: Extracting Interpretable Features from Claude 3 Sonnet},
       author={Templeton, Adly and Conerly, Tom and Marcus, Jonathan and Lindsey, Jack and Bricken, Trenton and Chen, Brian and Pearce, Adam and Citro, Craig and Ameisen, Emmanuel and Jones, Andy and Cunningham, Hoagy and Turner, Nicholas L and McDougall, Callum and MacDiarmid, Monte and Freeman, C. Daniel and Sumers, Theodore R. and Rees, Edward and Batson, Joshua and Jermyn, Adam and Carter, Shan and Olah, Chris and Henighan, Tom},
       year={2024},
       journal={Transformer Circuits Thread},
       url={https://transformer-circuits.pub/2024/scaling-monosemanticity/index.html}
    }

@misc{gpt2_attention_saes,
  author= {Connor Kissane and Robert Krzyzanowski and Arthur Conmy and Neel Nanda},
  url = {https://www.alignmentforum.org/posts/FSTRedtjuHa4Gfdbr},
  year = {2024},
  howpublished = {Alignment Forum},
  title = {Attention SAEs Scale to GPT-2 Small},
}

@inproceedings{
gao2025scaling,
title={Scaling and evaluating sparse autoencoders},
author={Leo Gao and Tom Dupre la Tour and Henk Tillman and Gabriel Goh and Rajan Troll and Alec Radford and Ilya Sutskever and Jan Leike and Jeffrey Wu},
booktitle={The Thirteenth International Conference on Learning Representations},
year={2025},
url={https://openreview.net/forum?id=tcsZt9ZNKD}
}

@inproceedings{
kantamneni2025sparse,
    title={Are Sparse Autoencoders Useful? A Case Study in Sparse Probing},
    author={Subhash Kantamneni and Joshua Engels and Senthooran Rajamanoharan and Max Tegmark and Neel Nanda},
    booktitle={Forty-second International Conference on Machine Learning},
    year={2025},
    url={https://openreview.net/forum?id=rNfzT8YkgO}
}

@inproceedings{tenney2019bert,
  title={BERT Rediscovers the Classical NLP Pipeline},
  author={Ian Tenney and Dipanjan Das and Ellie Pavlick},
  booktitle={Annual Meeting of the Association for Computational Linguistics},
  year={2019},
  url={https://api.semanticscholar.org/CorpusID:155092004}
}

@inproceedings{hewitt2019structural,
  title={A Structural Probe for Finding Syntax in Word Representations},
  author={John Hewitt and Christopher D. Manning},
  booktitle={North American Chapter of the Association for Computational Linguistics},
  year={2019},
  url={https://api.semanticscholar.org/CorpusID:106402715}
}

@article{kramar2024atp,
  title={Atp*: An efficient and scalable method for localizing LLM behaviour to components},
  author={Kram{\'a}r, J{\'a}nos and Lieberum, Tom and Shah, Rohin and Nanda, Neel},
  journal={arXiv preprint arXiv:2403.00745},
  year={2024}
}

@inproceedings{
dunefsky2024transcoders,
title={Transcoders find interpretable {LLM} feature circuits},
author={Jacob Dunefsky and Philippe Chlenski and Neel Nanda},
booktitle={The Thirty-eighth Annual Conference on Neural Information Processing Systems},
year={2024},
url={https://openreview.net/forum?id=J6zHcScAo0}
}

@article{ameisen2025circuit,
  author={Ameisen, Emmanuel and Lindsey, Jack and Pearce, Adam and Gurnee, Wes and Turner, Nicholas L. and Chen, Brian and Citro, Craig and Abrahams, David and Carter, Shan and Hosmer, Basil and Marcus, Jonathan and Sklar, Michael and Templeton, Adly and Bricken, Trenton and McDougall, Callum and Cunningham, Hoagy and Henighan, Thomas and Jermyn, Adam and Jones, Andy and Persic, Andrew and Qi, Zhenyi and Ben Thompson, T. and Zimmerman, Sam and Rivoire, Kelley and Conerly, Thomas and Olah, Chris and Batson, Joshua},
  title={Circuit Tracing: Revealing Computational Graphs in Language Models},
  journal={Transformer Circuits Thread},
  year={2025},
  url={https://transformer-circuits.pub/2025/attribution-graphs/methods.html}
}

@inproceedings{caucheteux2021disentangling,
  title={Disentangling syntax and semantics in the brain with deep networks},
  author={Caucheteux, Charlotte and Gramfort, Alexandre and King, Jean-Remi},
  booktitle={International conference on machine learning},
  pages={1336--1348},
  year={2021},
  organization={PMLR}
}

@inproceedings{sundararajan2017axiomatic,
  title={Axiomatic attribution for deep networks},
  author={Sundararajan, Mukund and Taly, Ankur and Yan, Qiqi},
  booktitle={International conference on machine learning},
  pages={3319--3328},
  year={2017},
  organization={PMLR}
}

@article{goldowskydill2023localizing,
  author  = {Goldowsky-Dill, Nicholas and MacLeod, Chris and Sato, Lucas Jun Koba and Arora, Aryaman},
  title   = {Localizing Model Behavior with Path Patching},
  journal = {CoRR},
  volume  = {abs/2304.05969},
  year    = {2023},
  url     = {https://arxiv.org/abs/2304.05969},
}

@inproceedings{Stolfo2023AMI,
  title={A Mechanistic Interpretation of Arithmetic Reasoning in Language Models using Causal Mediation Analysis},
  author={Alessandro Stolfo and Yonatan Belinkov and Mrinmaya Sachan},
  booktitle={Conference on Empirical Methods in Natural Language Processing},
  year={2023},
  url={https://api.semanticscholar.org/CorpusID:258865170}
}

@inproceedings{muller-eberstein-etal-2023-subspace,
    title = "Subspace Chronicles: How Linguistic Information Emerges, Shifts and Interacts during Language Model Training",
    author = {M{\"u}ller-Eberstein, Max  and
      van der Goot, Rob  and
      Plank, Barbara  and
      Titov, Ivan},
    editor = "Bouamor, Houda  and
      Pino, Juan  and
      Bali, Kalika",
    booktitle = "Findings of the Association for Computational Linguistics: EMNLP 2023",
    month = dec,
    year = "2023",
    address = "Singapore",
    publisher = "Association for Computational Linguistics",
    url = "https://aclanthology.org/2023.findings-emnlp.879/",
    doi = "10.18653/v1/2023.findings-emnlp.879",
    pages = "13190--13208",
}

@inproceedings{NEURIPS2024_47c7edad,
 author = {Tigges, Curt and Hanna, Michael and Yu, Qinan and Biderman, Stella},
 booktitle = {Advances in Neural Information Processing Systems},
 editor = {A. Globerson and L. Mackey and D. Belgrave and A. Fan and U. Paquet and J. Tomczak and C. Zhang},
 pages = {40699--40731},
 publisher = {Curran Associates, Inc.},
 title = {LLM Circuit Analyses Are Consistent Across Training and Scale},
 url = {https://proceedings.neurips.cc/paper_files/paper/2024/file/47c7edadfee365b394b2a3bd416048da-Paper-Conference.pdf},
 volume = {37},
 year = {2024}
}

@article{PropBank,
author = {Palmer, Martha and Gildea, Daniel and Kingsbury, Paul},
title = {The Proposition Bank: An Annotated Corpus of Semantic Roles},
year = {2005},
issue_date = {March 2005},
publisher = {MIT Press},
address = {Cambridge, MA, USA},
volume = {31},
number = {1},
issn = {0891-2017},
url = {https://doi.org/10.1162/0891201053630264},
doi = {10.1162/0891201053630264},
journal = {Comput. Linguist.},
month = mar,
pages = {71–106},
numpages = {36}
}

@article{fillmore1976frame,
  title={Frame semantics and the nature of language},
  author={Fillmore, Charles J},
  journal={Annals of the New York Academy of Sciences},
  volume={280},
  number={1},
  pages={20--32},
  year={1976},
  publisher={Wiley Online Library}
}

@inproceedings{FrameNet,
    title = "The {B}erkeley {F}rame{N}et Project",
    author = "Baker, Collin F.  and
      Fillmore, Charles J.  and
      Lowe, John B.",
    booktitle = "36th Annual Meeting of the Association for Computational Linguistics and 17th International Conference on Computational Linguistics, Volume 1",
    month = aug,
    year = "1998",
    address = "Montreal, Quebec, Canada",
    publisher = "Association for Computational Linguistics",
    url = "https://aclanthology.org/P98-1013/",
    doi = "10.3115/980845.980860",
    pages = "86--90"
}

@inproceedings{
mueller2025mib,
title={{MIB}: A Mechanistic Interpretability Benchmark},
author={Aaron Mueller and Atticus Geiger and Sarah Wiegreffe and Dana Arad and Iv{\'a}n Arcuschin and Adam Belfki and Yik Siu Chan and Jaden Fried Fiotto-Kaufman and Tal Haklay and Michael Hanna and Jing Huang and Rohan Gupta and Yaniv Nikankin and Hadas Orgad and Nikhil Prakash and Anja Reusch and Aruna Sankaranarayanan and Shun Shao and Alessandro Stolfo and Martin Tutek and Amir Zur and David Bau and Yonatan Belinkov},
booktitle={Forty-second International Conference on Machine Learning},
year={2025},
url={https://openreview.net/forum?id=sSrOwve6vb}
}

@article{wolf2019huggingface,
  title={Huggingface's transformers: State-of-the-art natural language processing},
  author = {Thomas Wolf and Lysandre Debut and Victor Sanh and Julien Chaumond and Clement Delangue and Anthony Moi and Pierric Cistac and Tim Rault and R{\'e}mi Louf and Morgan Funtowicz and Joe Davison and Sam Shleifer and Patrick von Platen and Clara Ma and Yacine Jernite and Julien Plu and Canwen Xu and Teven Le Scao and Sylvain Gugger and Mariama Drame and Quentin Lhoest and Alexander Rush},
  journal={arXiv preprint arXiv:1910.03771},
  year={2019}
}

@article{touvron2023llama,
  title={Llama: Open and efficient foundation language models},
  author={Touvron, Hugo and Lavril, Thibaut and Izacard, Gautier and Martinet, Xavier and Lachaux, Marie-Anne and Lacroix, Timoth{\'e}e and Rozi{\`e}re, Baptiste and Goyal, Naman and Hambro, Eric and Azhar, Faisal and others},
  journal={arXiv preprint arXiv:2302.13971},
  year={2023}
}

@inproceedings{kim2025reasoning,
  title={Reasoning circuits in language models: A mechanistic interpretation of syllogistic inference},
  author={Kim, Geonhee and Valentino, Marco and Freitas, Andr{\'e}},
  booktitle={Findings of the Association for Computational Linguistics: ACL 2025},
  pages={10074--10095},
  year={2025}
}

@article{he2024learning,
  title={Learning to grok: Emergence of in-context learning and skill composition in modular arithmetic tasks},
  author={He, Tianyu and Doshi, Darshil and Das, Aritra and Gromov, Andrey},
  journal={Advances in Neural Information Processing Systems},
  volume={37},
  pages={13244--13273},
  year={2024}
}

@misc{santorini2007syntax,
  author = {Santorini, Beatrice and Kroch, Anthony},
  title = {The Syntax of Natural Language: An Online Introduction},
  year = {2007},
  url = {https://www.ling.upenn.edu/~beatrice/syntax-textbook},
  urldate = {2024-11-20},
}

@Book{jurafsky2009speech,
  author =       "Daniel Jurafsky and James H. Martin",
  title =        "Speech and Language Processing: An Introduction to
                 Natural Language Processing, Computational Linguistics,
                 and Speech Recognition with Language Models",
  year =         "2025",
  url = {https://web.stanford.edu/~jurafsky/slp3/},
  note = "Online manuscript released August 24, 2025",
  edition =         "3rd",
  }

@inproceedings{aljaafari-etal-2025-carma,
    title = "{CARMA}: Enhanced Compositionality in {LLM}s via Advanced Regularisation and Mutual Information Alignment",
    author = "Aljaafari, Nura  and
      Carvalho, Danilo  and
      Freitas, Andre",
    editor = "Christodoulopoulos, Christos  and
      Chakraborty, Tanmoy  and
      Rose, Carolyn  and
      Peng, Violet",
    booktitle = "Proceedings of the 2025 Conference on Empirical Methods in Natural Language Processing",
    month = nov,
    year = "2025",
    address = "Suzhou, China",
    publisher = "Association for Computational Linguistics",
    url = "https://aclanthology.org/2025.emnlp-main.822/",
    doi = "10.18653/v1/2025.emnlp-main.822",
    pages = "16250--16270",
    ISBN = "979-8-89176-332-6",
}

@book{carnie2021syntax,
  title={Syntax: A generative introduction},
  author={Carnie, Andrew},
  year={2021},
  publisher={John Wiley \& Sons}
}

@article{pile,
  title={The {P}ile: An 800GB Dataset of Diverse Text for Language Modeling},
  author={Gao, Leo and Biderman, Stella and Black, Sid and Golding, Laurence and Hoppe, Travis and Foster, Charles and Phang, Jason and He, Horace and Thite, Anish and Nabeshima, Noa and Presser, Shawn and Leahy, Connor},
  journal={arXiv preprint arXiv:2101.00027},
  year={2020}
}

\appendix
\section{Linguistic Background: Predicate–Argument Structure and Thematic Roles}
\label{app:linguistic-background}

\subsection{Predicate–Argument Structure}
Predicates, typically verbs, denote events or states and introduce \emph{argument positions} corresponding to event participants \citep{santorini2007syntax}. In \textit{``The courier delivered the package to the office with the truck''}, the predicate \textit{deliver} evokes a transfer event with participants such as the actor (\textit{courier}), the transferred entity (\textit{package}), the destination (\textit{office}), and the means (\textit{truck}). These participants may be realised as subjects, objects, or prepositional phrases, but surface form does not uniquely determine semantic function:  ``with X" may express \textsc{Instrument} (\textit{with the truck}) or \textsc{Comitative} (\textit{with Mary}), and the same role may occur in different syntactic configurations (e.g., dative alternation: \textit{``give the book to Mary''} vs.\ \textit{``give Mary the book''}).

\subsection{Thematic Roles ($\theta$-Roles)}
\textbf{Thematic roles} provide a shallow semantic representation that captures how participants relate to an event \citep{fillmore1976frame}.  
Roles such as \textsc{Agent}, \textsc{Theme}, \textsc{Goal}, \textsc{Instrument}, \textsc{Location}, \textsc{Time}, and \textsc{Beneficiary} support abstraction across syntactic frames and lexical variation.  
This level of representation identifies \emph{who did what to whom, where, and when} without committing to deeper logical structure \citep{jurafsky2009speech}. Shallow semantics refers to the intermediate layer commonly modelled in semantic role labelling \citep{jurafsky2009speech}.

\subsection{Predicate–Argument Binding}
\textbf{Predicate–argument binding} is the process of assigning thematic roles to the appropriate argument tokens. It forms the structured substrate on which more complex semantic composition builds: without correct role assignment, higher-level interpretation (e.g., quantification, scope, or discourse reference) cannot proceed.  
Our study isolates this binding mechanism, focusing on shallow semantics, and does not address deeper semantic phenomena such as quantifier scope or anaphora.

\subsection{Computational Instantiation in Transformers} In transformer models, predicate–argument binding might arise through:  
(i) attention heads routing information between predicates and their arguments,  (ii) distributed representations encoding role–filler associations, or  (iii) localised circuits whose coordinated activity is \emph{causally necessary} for role prediction. Our study tests for such circuits by combining causal edge-attribution patching, temporal emergence analysis, and cross-model comparison.  
Our goal is to determine whether transformers encode thematic roles via \emph{computationally indispensable} mechanisms that are structurally localised, temporally trackable, and partially conserved across architectures.

\paragraph{Relevance for interpretability.} If transformers encode predicate–argument binding via compact circuits, this suggests that meaningful semantic abstractions emerge naturally during training. Such findings provide mechanistic links between representation learning and the acquisition of linguistic structure, and offer principled targets for editing semantic behaviour.

\section{Dataset Generation Formalisation and Intervention Specifications}
\label{app:data}

\subsection{Task: Role-Cross Next-Token Prediction}
We construct \textbf{role-cross minimal pairs} to isolate the semantic role processing. Each pair consists of two incomplete prompts that differ only in their role-indicating scaffold:

\begin{equation}
\begin{aligned}
x^{(r)} &= \text{``The agent verb theme scaffold}^{(r)}\text{''} \\
x^{(s)} &= \text{``The agent verb theme scaffold}^{(s)}\text{''}
\end{aligned}
\end{equation}

where scaffold$^{(r)}$ (e.g., ``to the'', ``about the'') indicates the 
semantic role $r$ of the next token. The two contexts are constructed such that:
\begin{itemize}[leftmargin=*, itemsep=1pt]
    \item In clean context $x^{(r)}$, target token $y^{(r)}$ is role-appropriate 
          and most probable
    \item In corrupted context $x^{(s)}$, a \textbf{different} token $y^{(s)}$ 
          appropriate for role $s$ should be most probable
    \item Both $y^{(r)}$ and $y^{(s)}$ are drawn from a cross-role lexicon, 
          ensuring they are valid fillers for their respective roles, but not 
          for each other's roles
\end{itemize}

\paragraph{Evaluation Task.}
The model performs role-cross prediction correctly if it predicts the 
role-appropriate target in each context:
\begin{equation}
\begin{split}
    \text{Accuracy} = \mathds{1}[y^{(r)} = \argmax_v P_\theta(v \mid x^{(r)})] \\
\land \mathds{1}[y^{(s)} = \argmax_v P_\theta(v \mid x^{(s)})]
\end{split}
\end{equation}

This measures whether the model correctly binds different role fillers based 
solely on the role scaffold, holding agent, verb, and theme constant.

\paragraph{Example.}
\begin{align*}
x^{(\textsc{Goal})} &= \text{``The driver sent the wall to the''} \\
    \quad &\to y^{(\textsc{Goal})} = \text{``office''} \\
x^{(\textsc{Topic})} &= \text{``The driver sent the wall about the''} \\
    \quad &\to y^{(\textsc{Topic})} = \text{``plan''}
\end{align*}

The scaffolds ``to the'' and ``about the'' have the same token length (parity), 
but should activate different role-specific vocabularies.

\subsection{Frame-Based Template Construction}
We construct role-cross pairs inspired by PropBank \citep{PropBank} and FrameNet \citep{FrameNet} annotations using frame-based templates. Each template consists of:
\begin{itemize}[leftmargin=*, itemsep=1pt]
    \item \textbf{Frame}: Semantic structure (e.g., \textsc{Transfer}, 
          \textsc{Communication})
    \item \textbf{Verb}: Single-token predicate (e.g., ``sent'', ``prepared'')
    \item \textbf{Scaffold}: Role-indicating preposition phrase (e.g., ``to 
          the'' for \textsc{Goal}, ``about the'' for \textsc{Topic})
    \item \textbf{Agent}: Single-token subject (e.g., ``driver'', ``worker'')
    \item \textbf{Theme}: Single-token object (e.g., ``wall'', ``package'')
    \item \textbf{Target}: Role-specific single-token filler (e.g., ``office'' 
          for \textsc{Goal}, ``plan'' for \textsc{Topic})
\end{itemize}

\paragraph{Single-Token Constraint.}
All lexical items must tokenise to exactly one token when preceded by a space (GPT-NeoX/Llama convention). We validate using the target model's tokeniser and filter out multi-token words. This ensures: (i) precise position alignment 
for activation patching, and (ii) unambiguous attribution to specific lexical 
items.

\paragraph{Token Parity Enforcement.}
For each role-cross pair $(x^{(r)}, x^{(s)})$, we enforce \textbf{strict 
token-level parity}: $|\text{toks}(x^{(r)})| = |\text{toks}(x^{(s)})|$. This 
is achieved by:
\begin{enumerate}[leftmargin=*, itemsep=1pt]
    \item Grouping scaffolds by token length (e.g., 2-token: ``to the'', ``in 
          the'', ``about the''; 1-token: ``at'', ``on'')
    \item Only pairing roles whose scaffolds have matching token lengths
    \item Keeping agent, verb, and theme constant across pairs
    \item Rejecting pairs that violate parity after substitution
\end{enumerate}

Token parity is \textbf{essential for EAP-IG}, as activation differences 
$\Delta_u = \mathbf{z}_u^{(r)} - \mathbf{z}_u^{(s)}$ require position-aligned representations.

\paragraph{Role-Specific Lexicons.}
Each semantic role has a curated lexicon of plausible fillers:
\begin{itemize}[leftmargin=*, itemsep=1pt]
    \item \textbf{Goal}: Places and people (``office'', ``student'', 
          ``school'')
    \item \textbf{Location}: Places (``kitchen'', ``office'', ``park'')
    \item \textbf{Instrument}: Tools (``hammer'', ``knife'', ``drill'')
    \item \textbf{Material}: Substances (``steel'', ``wood'', ``stone'')
    \item \textbf{Topic}: Abstract concepts (``plan'', ``idea'', ``issue'')
    \item \textbf{Beneficiary}: People (``student'', ``client'', ``friend'')
\end{itemize}

Lexicons are designed such that tokens are \textbf{role-discriminative}: strongly preferred in their primary role but less probable in other roles (e.g., ``hammer'' is a good \textsc{Instrument} but bad \textsc{Topic}).

\subsection{Generation Procedure}

For each target role $r$ and desired sample size $N$:
\begin{enumerate}[leftmargin=*, itemsep=1pt]
    \item Sample target token $y^{(r)}$ from role $r$'s lexicon
    \item Identify corrupt role $s \neq r$ such that:
          \begin{itemize}
              \item Scaffolds for $r$ and $s$ have matching token lengths 
                    (parity constraint)
              \item Role $s$ has a distinct lexicon (ensures different target 
                    $y^{(s)}$)
          \end{itemize}
    \item Sample verb-agent-theme triple compatible with both roles
    \item Construct clean prefix $x^{(r)}$ with scaffold$^{(r)}$
    \item Construct corrupted prefix $x^{(s)}$ by replacing scaffold$^{(r)}$ 
          with scaffold$^{(s)}$
    \item Sample foil token $y^{(s)}$ from role $s$'s lexicon
    \item Validate:
          \begin{itemize}
              \item Token parity: $|\text{toks}(x^{(r)})| = 
                    |\text{toks}(x^{(s)})|$
              \item No leakage: Neither $y^{(r)}$ nor $y^{(s)}$ appears in 
                    prefixes
              \item All tokens are single-token
          \end{itemize}
    \item If validation passes, add $(x^{(r)}, x^{(s)}, y^{(r)}, y^{(s)})$ to 
          dataset
    \item Repeat until $N$ valid pairs obtained (patience limit: $30N$ attempts)
\end{enumerate}

\subsection{Filtering Procedure}

We filter generated pairs to retain only examples where the model predicts the role-appropriate target in \textbf{both} contexts. This ensures circuits are functionally active: the model successfully performs role-specific binding in both clean and corrupted contexts. Examples where either prediction is incorrect are discarded, as they would not reflect active role processing.


\subsection{Dataset Statistics}

All data is in English and after filtering with models, we obtain:
\begin{itemize}[leftmargin=*, itemsep=1pt]
    \item \textbf{Token parity}: 100\% (all pairs have matching token counts)
    \item \textbf{Dual prediction accuracy}: 100\% (by construction, 
          post-filtering: model predicts the correct target in both clean and 
          corrupted contexts)
    \item \textbf{Cross-role coverage}: Each role has at least ${\geq}450$ examples
\end{itemize}

Table~\ref{tab:dataset_summary} provides a per-role and model breakdown.
\begin{table*}[h]
\centering

\small
\begin{tabular}{lcccc}
\toprule
\textbf{Examples per Role} & \emph{Pythia-14M} & \emph{Pythia-410M} & \emph{Pythia-1B} & \emph{LLaMA-1B} \\
\midrule
{Goal}        & 845 & 895 & 1052 & 491 \\
{Location}    & 904 & 673 & 975 & 815 \\
{Source}      & 667 & 1206 & 505 & 612 \\
{Path}        & 802 & 894 & 707 & 959 \\
{Instrument}  & 459 & 1460 & 1212 & 1098 \\
{Beneficiary} & 773 & 837 & 621 & 1502 \\
{Topic}       & 1120 & 1323 & 491 & 1179 \\
{Time}        & 557 & 700 & 465 & 712 \\
\bottomrule
\end{tabular}
\caption{Role-cross dataset statistics after filtering for all models. All examples satisfy: (i) strict token parity between clean and corrupted contexts, (ii) model predicts target correctly in both contexts.}\label{tab:dataset_summary}
\end{table*}

\section{Computation and parameters specifications}
\subsection{Hyperparameter Selection}\label{app:hyperparameters}
All hyperparameters were selected based on circuit size constraints, 
computational feasibility, and robustness to measurement noise. We report 
choices for attribution, sparsity measurement, emergence detection, and 
cross-scale comparison.

\paragraph{Attribution (EAP-IG).}
\begin{itemize}[leftmargin=*, itemsep=2pt]
    \item \textbf{Integrated gradients steps}: 5. We use only 5 steps due to the large number of training checkpoints analysed. Ablation tests (not shown) confirmed 5 steps provide stable attributions whilst maintaining computational tractability for multi-checkpoint analysis.
    \item \textbf{Metric}: Negative log-probability of the role-appropriate 
          target token in the clean context: 
          $\mathcal{L} = -\log P_\theta(y^{(r)} \mid x^{(r)})$. This loss 
          quantifies the model's confidence in correct role binding.
\end{itemize}

\paragraph{Sparsity and Localization (RQ1).}
\begin{itemize}[leftmargin=*, itemsep=2pt]
    \item \textbf{Top-K node mass}: $K \in \{5, 10, 20\}$. We report the 
          fraction of total attribution mass captured by the top-$K$ 
          highest-mass nodes. These values span from highly concentrated cores 
          ($K{=}5$) to broader component sets ($K{=}20$).
    \item \textbf{Gini coefficient}: Computed over all in-circuit node masses 
          (nodes with non-zero attribution). Higher Gini indicates more unequal 
          mass distribution (tighter concentration).
    \item \textbf{Rationale}: All discovered circuits contain fewer than 40 
          active nodes at convergence, making $K{=}20$ a natural threshold 
          capturing approximately 50\% of the component space whilst focusing 
          on high-importance nodes.
\end{itemize}

\paragraph{Emergence Dynamics (RQ2).}
\begin{itemize}[leftmargin=*, itemsep=2pt]
    \item \textbf{Indispensability threshold}: $M(\mathcal{C}_t) - 
          M(\mathcal{E}_t) < 0$ for $\geq 2$ consecutive checkpoints. Circuit 
          performance must fall persistently below baseline to avoid transient 
          noise.
    \item \textbf{Change-point detection}: Two-segment piecewise linear 
          regression applied to faithfulness and Top-20 mass trajectories. 
          Bootstrap resampling ($n{=}1{,}000$) estimates 95\% confidence 
          intervals. Minimum segment length: 2 checkpoints (prevents overfitting 
          to single-step noise).
    \item \textbf{Consolidation criterion}: Jaccard overlap $\geq 0.6$ between 
          top-$K{=}20$ node sets over a 2-step sliding window. This threshold 
          balances sensitivity (detects stabilisation) against noise (ignores 
          minor fluctuations).
\end{itemize}

\paragraph{Cross-Scale Comparison (RQ3).}
\begin{itemize}[leftmargin=*, itemsep=2pt]
    \item \textbf{Node/edge overlap}: Top-$K{=}30$ components by absolute 
          attribution mass. We increase $K$ from 20 (used in RQ1/RQ2) to 30 for 
          cross-scale comparison because larger models (410M, 1B) have more 
          active nodes; $K{=}30$ ensures we compare substantive component sets 
          whilst maintaining focus on high-importance nodes.
    \item \textbf{Spectral distance}: Computed on top-$K{=}50$ edges using the 
          lowest $k{=}20$ Laplacian eigenvalues. We use more edges ($K{=}50$) 
          for spectral analysis than overlap ($K{=}30$) because eigenvalue 
          computation requires sufficient connectivity to yield stable spectra. 
          The first 20 eigenvalues capture low-frequency flow structure whilst 
          remaining computationally tractable.
    \item \textbf{Rationale}: Since all circuits contain $<$40 active nodes at 
          convergence, $K{=}30$ captures ${\sim}$75\% of the component space. 
          This "hard 50\% rule" ensures overlap metrics reflect substantive 
          similarity rather than trivial peripheral agreement. For edges, 
          $K{=}50$ balances spectral stability with focus on high-attribution 
          connections.
\end{itemize}

\paragraph{Data Filtering.}
\begin{itemize}[leftmargin=*, itemsep=2pt]
    \item \textbf{Correctness criterion}: Retain only examples where the model 
          predicts the role-appropriate target correctly in \textbf{both} clean 
          and corrupted contexts at baseline (full model). Formally: 
          $y^{(r)} = \argmax_v P_\theta(v \mid x^{(r)})$ AND 
          $y^{(s)} = \argmax_v P_\theta(v \mid x^{(s)})$.
    \item \textbf{Rationale}: This dual-correctness filter ensures discovered 
          circuits are functionally active---the model successfully performs 
          role binding in both contexts, guaranteeing circuits supporting this 
          capability are present and engaged. Analysing only correctly predicted examples is standard practice in mechanistic interpretability 
          \citep{wang2022interpretability, conmy2023towards} as it isolates 
          functional mechanisms rather than failure modes.
\end{itemize}

\subsection{Computational Constraints.}\label{app:software}
All experiments were conducted on an NVIDIA RTX A6000 GPU. Attribution over multiple checkpoints required approximately 5 min per role for \emph{Pythia-1b}. The choice of 5 IG steps was necessary to complete the temporal analysis within feasible compute budgets whilst maintaining attribution stability, as confirmed by spot-checks with higher step counts on selected checkpoints.

\subsection{Software Specification.}
Experiments were conducted using Python~3.11.13 with NumPy 1.26.4, scikit-learn~1.7.0, scipy~1.15.3, seaborn~0.13.2, tokenizers~0.21.1, torch~2.7.1, transformer-lens~2.16.1, and transformers~4.52.4, and trace~0.2.0.

\subsection{Models Specifications.} 
All models were obtained from Hugging Face \cite{wolf2019huggingface} and used under their respective intended use, following their respective licenses: Llama 3.2 (Meta Llama 3 Community), Pythia Models (Apache license 2.0), we summarise their key characteristics in Table \ref{tab:model_properties}. Pythia models were mainly pre-trained on English data. LLama, on the other hands, has additional multilingual capabilities, including, English, German, French, Italian, Portuguese, Hindi, Spanish, and Thai. The models employ the following tokenisation approaches: Llama 3.2 uses SentencePiece-based BPE, combining 100K tokens from Tiktoken3 with 28K additional tokens to enhance multilingual performance, while Pythia employs GPT-NeoX. 

\begin{table}[ht]
\centering
\renewcommand{\arraystretch}{1.2} 
\setlength{\tabcolsep}{6pt} 
\resizebox{\columnwidth}{!}{%
\begin{tabular}{|l|c|c|c|c|c|c|}
\hline
\textbf{Model} & \textbf{Parameters} & \textbf{Layers} & \textbf{D\textsubscript{model}} & \textbf{Heads} & \textbf{Activation}  \\
\hline
Pythia 14M  & 1.2M & 6  & 128  & 4  & gelu \\
Pythia 410M  & 302M & 24  & 2048  & 16  & gelu \\
Pythia 1B  & 805M & 16  & 2048  & 8  & gelu \\
LLaMA3.2 1B  & 1.1B & 16  & 2048  & 32  & SiLU \\
\hline
\end{tabular}%
}
\caption{Summary of model architectures. \textbf{Parameters} is the total number of trainable parameters; \textbf{Layers} is number of transformer layers; \textbf{D\textsubscript{model}}: size of word embeddings and hidden states; \textbf{Heads}: number of self-attention heads; and \textbf{Activation}: activation function used in feedforward layers.}
\label{tab:model_properties}
\end{table}


\section{EAP-IG formalisation}\label{app:eap_ig}
EAP-IG combines the causal faithfulness of activation patching \citep{meng2022locating, wang2022interpretability} with the path-sensitivity of integrated gradients \citep{sundararajan2017axiomatic}, addressing key limitations of alternative approaches. We detailed its steps below.

\subsection{Transformers as Acyclic Graphs}
We adopt the graph-theoretic representation from \citet{hanna2024have}, modelling transformer computation as a directed acyclic graph $\mathcal{G}=(\mathcal{V},\mathcal{E})$ at module$\times$position granularity. A node $u\in\mathcal{V}$ corresponds to a specific module output:
\begin{equation}
\begin{split}
     u\in\mathcal{V}
\;\;\Longleftrightarrow\;\;
u=\big(\mathrm{type},\,\ell,\,h,\,i\big),
\quad\\
\mathrm{type}\in\{\mathrm{AttnHead},\mathrm{MLP}\},
\end{split}
\end{equation}

where $\ell$ is layer index, $i$ is the token position in the sequence, and $h$ is head index for attention heads; $h=\varnothing$ for MLPs. The activation $\mathbf{z}_u\in\mathbb{R}^d$ is the contribution of that module to the residual stream at position $i$.

An edge $(u\!\to\!v)\in\mathcal{E}$ exists if $\mathbf{z}_u$ is linearly mixed into the pre-activation input $\mathbf{s}_v$ of node $v$ via the residual stream and layer normalisation.  
This yields a fine-grained graph where each edge corresponds to a specific causal pathway between modules.

\subsection{EAP-IG Scoring Procedure}
For each clean/corrupt pair, let $\mathbf{x}^{(c)}$ and $\mathbf{x}^{(r)}$ be the input embedding sequences for $x^{(c)}{1:t}$ and $x^{(r)}{1:t}$, respectively, and $\mathbf{z}^{(c)}_u \text{ be the activation of node }u\text{ under }x^{(c)}_{1:t},$ and $\mathbf{z}^{(r)}_u \text{ be activation of node }u\text{ under }x^{(r)}_{1:t}$. We compute causal edge attributions through the following protocol:

\begin{enumerate}
    \item \textbf{Cache activations:} Run the model on both versions to obtain $\mathbf{z}_u^{(c)}$ and $\mathbf{z}_u^{(r)}$ for all nodes $u$.
    \item \textbf{Compute source deltas:} $\Delta_u = \mathbf{z}_u^{(r)} - \mathbf{z}_u^{(c)}$.
    \item \textbf{Interpolate inputs:} Define the straight-line interpolation $\mathbf{x}(\alpha) = \mathbf{x}^{(r)} + \alpha(\mathbf{x}^{(c)} - \mathbf{x}^{(r)})$ for $\alpha \in [0,1]$.
    \item \textbf{Gradient sampling:} For $m$ evenly spaced $\alpha_k$ and destination node $v$, at each step $k$, run a forward pass with input $\mathbf{x}(\alpha_k)$ and compute the gradient of the loss with respect to $v$'s pre-activation input: $\mathbf{g}_v^{(k)} = \partial L(\mathbf{x}(\alpha_k)) / \partial \mathbf{s}_v \in \mathbb{R}^d$.
    \item \textbf{Integrate:} Average over $m$ to obtain the integrated gradient estimate $\bar{\mathbf{g}}_v \;=\; \frac{1}{m}\sum_{k=1}^m \mathbf{g}^{(k)}_v$.
    \item \textbf{Edge score:} $S_{u\to v}^{\mathrm{IG}} = \Delta_u^\top \bar{\mathbf{g}}_v \in \mathbb{R}$.
\end{enumerate}

Intuitively, $S^{\mathrm{IG}}_{u\to v}$ approximates the first-order change in the loss if the contribution of $u$ to $v$'s input were replaced by its corrupted counterpart, \emph{averaged} along the naturalistic path from corrupt to clean inputs. By integrating over $\alpha$, the estimate reduces sensitivity to local saturation at the clean point and captures non-linear response accumulated along the path. We selected $m=5$, as we saw diminishing returns on higher values, similar to the results in \citep{hanna2024have}.

\subsection{Score Normalisation and Aggregation.} To enable fair comparison across layers and modules with different activation scales, we report both raw and normalised attribution scores. The normalised score for edge $(u\!\to\!v)$ is:
\begin{equation}
\label{eq:normalized-score}
\tilde{S}^{\mathrm{IG}}_{u\to v}
=
\frac{\Delta_u^\top\,\bar{\mathbf{g}}_v}{\|\Delta_u\|_2\,\|\bar{\mathbf{g}}_v\|_2 + \varepsilon},
\quad \varepsilon=10^{-8}.
\end{equation}
Unless otherwise stated, aggregated statistics are computed on raw scores; normalization is used for scale-invariant concentration measures.

For role-specific analysis, we aggregate edge scores into role-layer heatmaps by summing absolute scores over edges sharing a destination module:
\begin{equation}
\begin{split}
    \mathsf{Importance}^{(r)}(\mathrm{layer}=\ell, \mathrm{head}=h)
=\\
\sum_{\substack{(u\to v)\in\mathcal{E}:\\ v=(\mathrm{AttnHead},\ell,h,\cdot)}}
\big|S^{\mathrm{IG}}_{u\to v}\big|,
\end{split}
\end{equation}

and analogously for MLPs.  

\subsection{Circuits Evaluation metrics}\label{app:eap_metrics}
\textbf{{Faithfulness:}} the proportion of the clean--corrupt discrimination preserved by $\mathcal{C}$: 
\begin{equation}
    \mathrm{Faithfulness}(\mathcal{C})
=
\frac{M(\mathcal{C}) - M(\emptyset)}{M(\mathcal{E}) - M(\emptyset)},
\end{equation}
where $M(\mathcal{C})$ is the metric in Section~\ref{sec:temporal} under $\mathcal{C}$, $M(\mathcal{E})$ under all edges patched, and $M(\emptyset)$ under none patched.

\textbf{Temporal Consistency.}
Jaccard stability of top-$K$ edge sets between checkpoints $t$ and $t'$:
\begin{equation}
    \mathrm{Stability}(\mathcal{C}_t, \mathcal{C}_{t'})
=
\frac{|\mathcal{C}_t \cap \mathcal{C}_{t'}|}{|\mathcal{C}_t \cup \mathcal{C}_{t'}|}.
\end{equation}
High stability indicates that once a role circuit emerges, it persists across training.


\textbf{Bootstrap CI for Change-point}
We estimate a split point $t$ in $(x\!=\!\text{step},\,y\!=\!\text{faithfulness})$ by minimising the summed residuals of two OLS lines on $[x_1,\dots,x_t]$ and $[x_{t+1},\dots,x_n]$ with a minimum segment length of 3. For a 95\% CI, we resample pairs $(x,y)$ with replacement, sort by $x$, re-fit the split, map the bootstrap split $x_t^\star$ back to the original grid, and take the 2.5/97.5 percentiles over 1{,}000 replicates. We report the point estimate $\hat t_c$ and the percentile CI over steps.

\section{Metric Definitions and Interpretation}
\label{sec:metrics}

\paragraph{Setup.}
All metrics are computed on the \emph{in-circuit} subgraph for each \emph{role} and \emph{training step}.
Let $G=(V,E)$ be a directed graph whose vertices $v\in V$ are components
(attention heads $a\ell.h$, MLPs $m\ell$, special nodes \texttt{input}/\texttt{logits}),
and whose edges $e=(u\!\to\!v)\in E$ carry an attribution score $s(e)\in\mathbb{R}$ and a type
$\tau(e)\in\{\mathrm{Q,K,V,Flow}\}$. Unless stated, strength uses absolute attribution $|s(e)|$.

\paragraph{Node mass.}
Incident absolute attribution:
\begin{equation}
  \begin{split}
\mathrm{mass}(v)\;=\;\sum_{e\in \mathrm{Inc}(v)} |s(e)|,\\
\qquad
\mathrm{Mass}(G)\;=\;\sum_{v\in V}\mathrm{mass}(v).
  \end{split}
  \end{equation}
\textit{Note:} $\mathrm{Mass}(G)=2\sum_{e\in E}|s(e)|$ since each edge contributes to two endpoints.
We also report \textbf{Total mass} as a proxy for role salience at a step.

\subsubsection*{Sparsity \& Targeting (per role $\times$ step)}
\begin{description}
  \item[Top-$K$ node-mass proportion.]
  \begin{equation}
  \begin{split}
      \mathrm{TopK}(K)\;=\;\frac{\sum_{i=1}^{K} m_{(i)}}{\sum_{i=1}^{|V|} m_{(i)}},\\
  \quad m_{(1)}\ge \cdots \ge m_{(|V|)}.
  \end{split}
  \end{equation}
  Range $[0,1]$; higher~$\Rightarrow$~stronger sparsity. We report $K\in\{5,10,20\}$.

  \item[Top-$P$ coverage.]
  Minimal $K$ such that $\mathrm{TopK}(K)\ge P$, for $P\in\{0.80,0.90,0.95\}$.
  Lower $K$ indicates higher sparsity.

  \item[Gini coefficient (node mass).]
  Standard Gini on nonnegative masses; range $[0,1]$ ($1=$ all mass on one node).
  Primary comparator for \textbf{sparse localisation} (RQ1).

  comparable across graphs of different sizes.
\end{description}
\begin{table*}[t]
\centering
\small
\begin{tabular}{lcll}
\toprule
\textbf{Metric} & \textbf{Range} & \textbf{High} & \textbf{Low} \\
\midrule
Top-$K$ mass & $[0,1]$ & concentrated circuit (RQ1) & diffuse attribution \\
Top-$P$ coverage ($K$) & $\mathbb{N}$ & few nodes capture $P$ (sparse) & many nodes needed \\
Gini (mass) & $[0,1]$ & strong sparsity (RQ1) & uniform mass \\
Density & $[0,1]$ & saturated links (post-$\hat t_c$) & sparse links \\
Reciprocity & $[0,1]$ & feedback motifs & feed-forward routing \\
Avg out-degree & $\mathbb{R}_{\ge 0}$ & broad fan-out & narrow fan-out \\
Avg weighted out-degree & $\mathbb{R}_{\ge 0}$ & strong influence spread & weak influence \\
Edge-type mix & simplex & routing vs.\ residual balance & --- \\
Bridges & $\mathbb{N}$ & bottlenecks (ablation targets) & redundancy \\
Layer span & $\mathbb{N}$ & deeper integration & shallow circuit \\
Avg betweenness & $[0,1]$ (norm.) & coordinator hubs & flat routing \\
Top-$K$ Jaccard (step) & $[0,1]$ & persistent circuit & unstable set \\
Cross-model Jaccard & $[0,1]$ & architectural consistency & family/scale drift \\
Spectral distance $d_{\text{spec}}$ & $\mathbb{R}_{\ge 0}$ & similar flow geometry & divergent geometry \\
\bottomrule
\end{tabular}
\caption{Interpretation guide for graph metrics, with expected high and low values meaning.}
\label{tab:interpretation_tab}
\end{table*}
\subsubsection*{Structural / Connectivity (per role $\times$ step)}
\begin{description}
  \item[Nodes, Edges.] $|V|$ and $|E|$ of the in-circuit graph.

  \item[Density.]
  We use Network's directed density:
  \begin{equation}
  \begin{split}
  \mathrm{density}(G)=\frac{|E|}{|V|(|V|-1)}\in[0,1],
  \end{split}
  \end{equation}
  assuming no self-loops. (\emph{Implementation:} \texttt{nx.density}.)

  \item[Reciprocity.]
  Fraction of directed edges participating in reciprocated pairs:
  \begin{equation}
  \begin{split}
  \mathrm{recip}(G)=\frac{L_{\leftrightarrow}}{L}.
  \end{split}
  \end{equation}

  \item[Average out-degree / weighted out-degree.]
  \begin{equation}
  \begin{split}
  \overline{\deg^+}=\tfrac{1}{|V|}\sum_{v}\deg^+(v),\\
  \quad
  \overline{\deg^+_{w}}=\tfrac{1}{|V|}\sum_{v}\sum_{(v\to u)\in E}|s(v\!\to\!u)|.
  \end{split}
  \end{equation}

  \item[Edge-type fractions.]
  \begin{equation}
  \begin{split}
  \mathrm{frac}_{T}=\frac{|\{e\in E:\tau(e)=T\}|}{|E|},\\\quad T\in\{\mathrm{Q,K,V,Flow}\},\\\ \sum_T \mathrm{frac}_T=1.
  \end{split}
  \end{equation}

  \item[Bridges (undirected projection).]
  Count edges whose removal disconnects the \emph{undirected} projection of $G$ (structural bottlenecks).

  \item[Layer span.]
    
  \begin{equation}
  \begin{split}
  \mathrm{layer\_span}\;=\;\max_{v\in V}\mathrm{layer}(v)-\min_{v\in V}\mathrm{layer}(v).
  \end{split}
  \end{equation}

  \item[Average betweenness centrality.]
  $\overline{C_B}=\tfrac{1}{|V|}\sum_{v} C_B(v)$ on the directed graph (normalized; may use sampling for efficiency).
\end{description}

\subsubsection*{Emergence \& Stability (per role)}
\begin{description}
  \item[Detectability $t_{\text{det}}$ (optional).]
  First step where faithfulness exceeds an early-phase baseline plus $2\sigma$
  (baseline/variance from the first $2$ checkpoints), persisting for $\geq2$ checkpoints.

  \item[Indispensability $t_{\text{ind}}$.]
  Earliest step where ablating the discovered circuit yields a statistically significant performance drop
  that \emph{persists} for at least $2$ subsequent checkpoints.

  \item[Change-point $\hat t_c$ (with bootstrap CI).]
  Two-segment least-squares on $y_t\in\{\text{faithfulness},\mathrm{TopK}(K)\}$ with a minimum segment length of $3$.
  We report $\hat t_c$ with a nonparametric bootstrap $95\%$ CI.

  \item[Consolidation $t_{\text{cons}}$.]
  Earliest step post-$\hat t_c$ where Top-$K$ node sets stabilise:
  Jaccard$(V^{(K)}_t,V^{(K)}_{t'})\ge 0.6$ for a $3$-step window (persistence $=2$), with $K{=}20$.
\end{description}

\subsubsection*{Cross-scale / Cross-role Similarity}
\begin{description}
  \item[Node/edge overlap across models.]
  Mean Jaccard of Top-$K$ node/edge sets between two models, averaged over common checkpoints.
  (\emph{Defaults:} $K{=}30$.)

  \item[Spectral similarity.]
  For symmetrised, weighted Laplacians ($w_{uv}=|s(u\!\leftrightarrow\!v)|$),
  \[
  d_{\text{spec}}(G_i,G_j)=\mathrm{RMSE}\big(\lambda_{1:k}(G_i),\lambda_{1:k}(G_j)\big),
  \]
  where $\lambda_{1:k}$ are the $k$ smallest eigenvalues. Lower is more similar.
  (\emph{Defaults:} build undirected graphs from the Top-$50$ edges by $|s|$, $k{=}20$.)
  
  \item[Within-model role overlap.]
  Jaccard of Top-$K$ nodes between roles at fixed steps (specialisation vs.\ shared scaffolding).
\end{description}

\subsubsection*{Computation Conventions}
\begin{itemize}
  \item Mass/strength metrics use absolute attributions $|s(e)|$; sign-sensitive analyses are reported separately.
  \item Type fractions follow RDF edge labels $\tau(e)\in\{\mathrm{Q,K,V,Flow}\}$.
  \item Density uses \texttt{nx.density} on directed graphs (no self-loops). Bridges are computed on the undirected projection.
\end{itemize}

\paragraph{Causal Flow Visualisation} The causal flow diagrams (e.g., Figure~\ref{fig:causal-flow-grid}) visualise the dominant information pathways within discovered circuits. We construct a directed graph $G = (V, E)$ where nodes $V$ represent model components (e.g., attention heads $a_{\ell,h}$, MLP layers $m_\ell$, and output logits) and edges $E$ represent causal attribution paths with weights $w_{uv}$ quantifying the contribution of component $u$ to component $v$. \textbf{Edge selection.} Rather than visualising the complete circuit graph, we filter to the top-$k$ edges by attribution magnitude $|w_{uv}|$ among those marked as in-circuit. This selective visualisation serves two purposes: (1) it highlights the \emph{dominant} computational pathways that account for the majority of causal effect, as circuits exhibit high concentration, and (2) it ensures interpretability, as dense graphs obscure rather than illuminate mechanistic structure. We use quantile-based thresholds (95th percentile by default) with a minimum edge count (12) to ensure sufficient context for interpretation. \textbf{Graph layout.} Nodes are positioned via multipartite layout by layer depth $\ell$, flowing left-to-right from inputs ($\ell=-1$) through transformer layers to logits ($\ell=L+1$). Edge attributes encode: width $\propto |w_{uv}|$ (attribution strength), colour by edge type (Query/Key/Value composition versus residual flow), and style (solid for positive attribution, dashed for negative suppression). This representation exposes: (i) critical computational pathways for each semantic role, (ii) layer-wise concentration of circuit activity, and (iii) coordination patterns between attention mechanisms versus direct residual connections.
\subsubsection*{Interpretation Cheat Sheet} See Table~\ref{tab:interpretation_tab}, with full method algorithm in~\ref{alg:compass}.

\paragraph{Default parameters used in the paper.}
Unless otherwise noted: consolidation uses $K{=}20$, Jaccard $\ge 0.6$, persistence $=2$; cross-model overlap uses $K{=}30$; spectral similarity uses Top-$50$ edges and $k{=}20$ eigenvalues.

\begin{algorithm}[h!]
\caption{COMPASS: Causal-Temporal Circuit Discovery}
\label{alg:compass}
\begin{algorithmic}[1]
\Require Model checkpoints $\{\theta_t\}_{t=0}^T$, role-cross dataset $D$, 
         role $r \in \mathcal{R}$, top-$K$ threshold
\Ensure Circuit $\mathcal{C}_t^{(r)}$ for each $t$; emergence times 
        $(t_{\text{ind}}, \hat{t}_c, t_{\text{cons}})$

\State \textbf{Phase 1: Causal Localisation (EAP-IG)}
\For{each checkpoint $t = 0, \ldots, T$}
    \State Compute CNP scores $\{\Delta_{\theta_t}(y; r, s)\}$ for all 
           $(x^{(r)}, x^{(s)}) \in D$
    \State Run EAP-IG to obtain edge attributions 
           $\{S_{u \to v}^{\text{IG}}\}_{e \in \mathcal{E}}$ 
           (Appendix~\ref{app:eap_ig})
    \State Normalise: 
           $\tilde{S}_{u \to v}^{\text{IG}} \gets 
           S_{u \to v}^{\text{IG}} / \sum_e |S_e^{\text{IG}}|$
    \State Extract circuit: 
           $\mathcal{C}_t^{(r)} \gets \text{TopK}(
           \{|\tilde{S}_{u \to v}^{\text{IG}}|\}, K)$
\EndFor

\State \textbf{Phase 2: Temporal Monitoring}
\For{each checkpoint $t = 0, \ldots, T$}
    \State Compute faithfulness $F_t$ via ablation (See Sec.~\ref{sec:temporal})
    \State Compute stability $S_t$ via Jaccard (See Sec.~\ref{sec:temporal})
    \State Compute structural metrics: Top-$K$ node mass, Gini coefficient
\EndFor

\State \textbf{Phase 3: Emergence Detection}
\State Detect indispensability: $t_{\text{ind}} \gets 
       \min\{t : M_t(\mathcal{E}) - M_t(\mathcal{C}_t) > \epsilon 
       \text{ for } \geq 2 \text{ steps}\}$
\State Estimate functional transition: $\hat{t}_c \gets 
       \argmax_t R^2(\text{PiecewiseLinear}(\{F_t\}))$ with bootstrap CIs
\State Detect consolidation: $t_{\text{cons}} \gets 
       \min\{t : S_t \geq 0.6 \text{ for } \geq 2 \text{ steps}\}$

\State \Return $\{\mathcal{C}_t^{(r)}\}_{t=0}^T$, 
               $(t_{\text{ind}}, \hat{t}_c, t_{\text{cons}})$
\end{algorithmic}
\end{algorithm}

\section{Full Results}\label{app:full-results}
We add the full results for \emph{Pythia-14m}, \emph{Pythia-410M} and \emph{LLaMA-1B}. We note that all experiments were repeated on five different random seeds, and the reported results are the averaged graphs per model. 

\subsection{RQ1: Full Localisation Results}\label{app:rq1-details}
This section provides the full localisation analyses for all semantic roles, model scales, and training checkpoints, complementing the representative results in the main text. For each role, we report (i) mass–concentration statistics, (ii) sparsity and coverage metrics, and (iii) stability of component sets over checkpoints. 

\paragraph{Mass concentration and sparsity.}
Table~\ref{tab:rq1-sparsity-full} reports the Top--$K$ mass ($K\in\{5,10,20\}$) for all roles and models at the final checkpoint (143K steps). Across all settings, Top--20 nodes capture between 83\% and 99\% of attribution mass, confirming that role circuits remain highly concentrated even in larger scales.

\begin{table}[t]
\centering
\resizebox{\columnwidth}{!}{%
\begin{tabular}{lcccccccccc}
\toprule
& \multicolumn{3}{c}{\textsc{Pythia-14M}} & \multicolumn{3}{c}{\textsc{Pythia-410M}} & \multicolumn{3}{c}{\textsc{LLaMA-1B}} \\
\cmidrule(lr){2-4} \cmidrule(lr){5-7} \cmidrule(lr){8-10}
Role & T-5 & T-10 & T-20 & T-5 & T-10 & T-20 & T-5 & T-10 & T-20 \\
\midrule
\textsc{Beneficiary} & 0.465 & 0.769 & \textbf{0.988} & 0.345&0.58&\textbf{0.906 }& 0.4	&0.657	&\textbf{0.939} \\
\textsc{Instrument}  & 0.445 & 0.596 & \textbf{0.833} & 0.445&0.596&\textbf{0.833} & 0.484&	0.697	&\textbf{0.895} \\
\textsc{Location}    & 0.368 & 0.557 & \textbf{0.841} & 0.368&0.557&\textbf{0.841} & 0.413&0.636&\textbf{0.882}\\
\textsc{Time}        & 0.501 & 0.713 & \textbf{0.985}& 0.445	&0.610&\textbf{0.852}& 0.443&0.677&\textbf{0.961}\\
\bottomrule
\end{tabular}%
}
\caption{Top-$K$ mass concentration at final checkpoint (143K steps). Values show fraction of total attribution mass captured by $K$ highest-mass nodes, demonstrating strong localisation across roles and model scales.}
\label{tab:rq1-sparsity-full}
\end{table}

\begin{table}[t]
\centering
\resizebox{\columnwidth}{!}{%
\begin{tabular}{lccc}
\toprule
Role & \textsc{Pythia-14M} & \textsc{Pythia-410M} & \textsc{LLaMA-1B} \\
\midrule
\textsc{Beneficiary} & 11 / 13 / 15 & 16 / 20 / 23 & 14 / 17 / 22 \\
\textsc{Instrument}  & 11 / 14 / 17 & 19 / 24 / 28 & 15 / 21 / 25 \\
\textsc{Location}    & 12 / 15 / 17 & 19 / 23 / 27 & 17 / 21 / 25 \\
\textsc{Time}        & 10 / 14 / 18 & 18 / 24 / 28 & 14 / 17 / 20 \\
\bottomrule
\end{tabular}%
}
\caption{\textbf{Minimal node count required for coverage at the final checkpoint (143K steps).} Entries report the smallest number of nodes ($k$) needed to capture 80\%, 90\%, and 95\% of total attribution mass for each role and model, where small number of nodes constitute the majority of mass.}
\label{tab:rq1-coverage-full}
\end{table}

\paragraph{Coverage: minimal node set sizes.}
Table~\ref{tab:rq1-coverage-full} reports the smallest $k$ achieving 80\%, 90\%, and 95\% mass. Across all roles and models, fewer than 30 nodes suffice for 95\% coverage, again confirming that circuits remain compact even in 1B-scale architectures.

\begin{figure}
    \centering
    \includegraphics[width=0.8\linewidth]{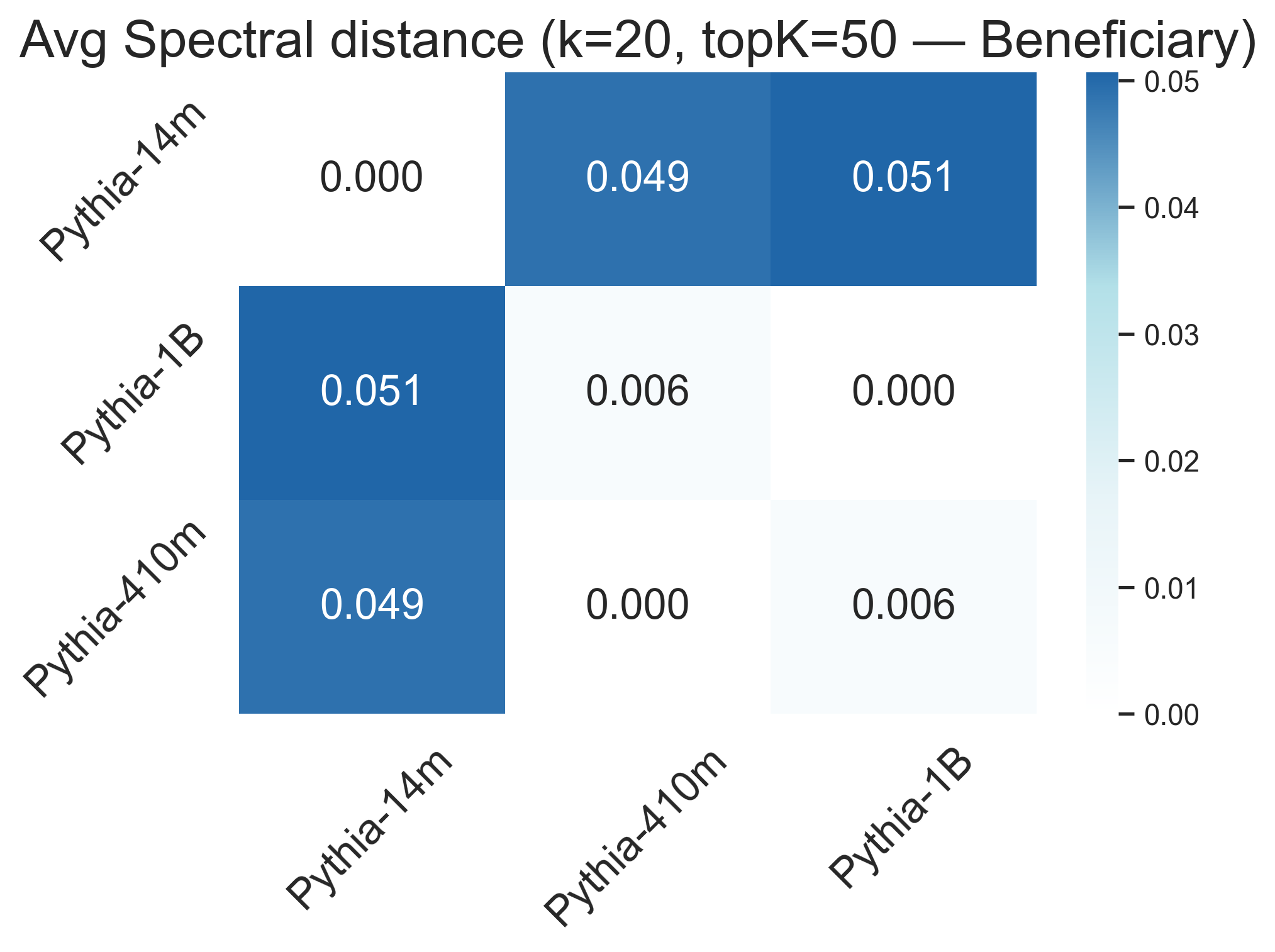}
    \includegraphics[width=0.8\linewidth]{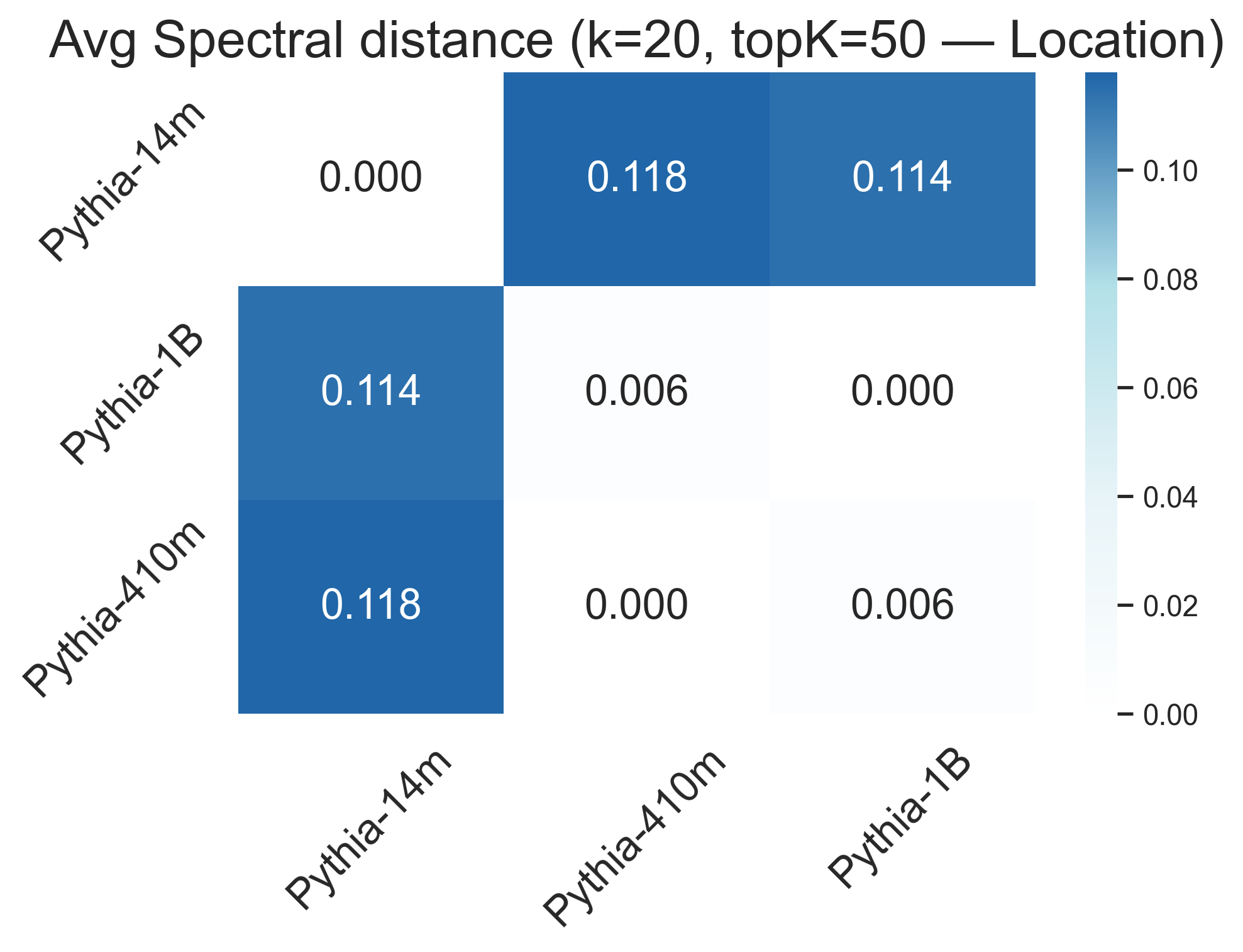}
    \includegraphics[width=0.8\linewidth]{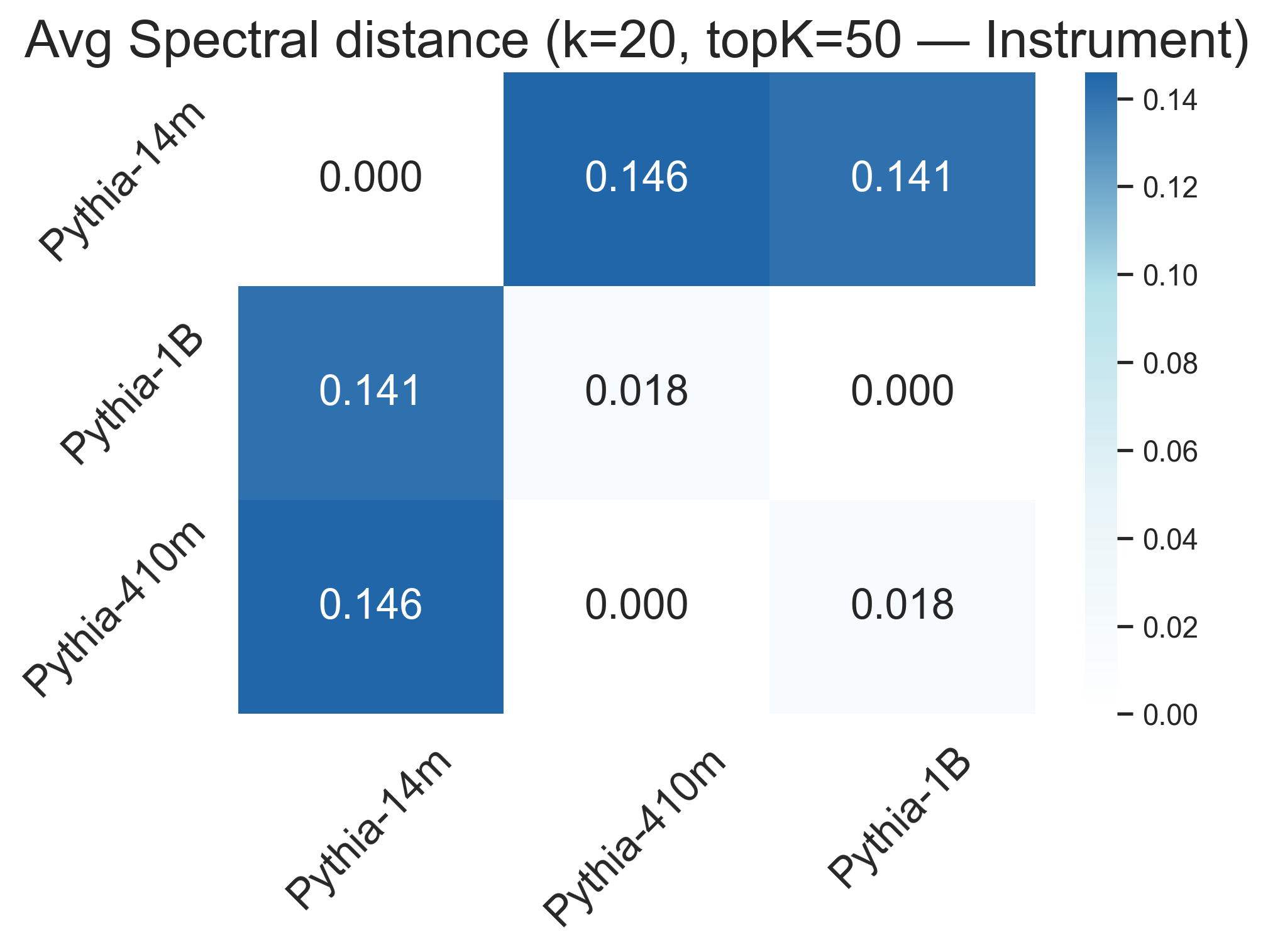}
    \includegraphics[width=0.8\linewidth]{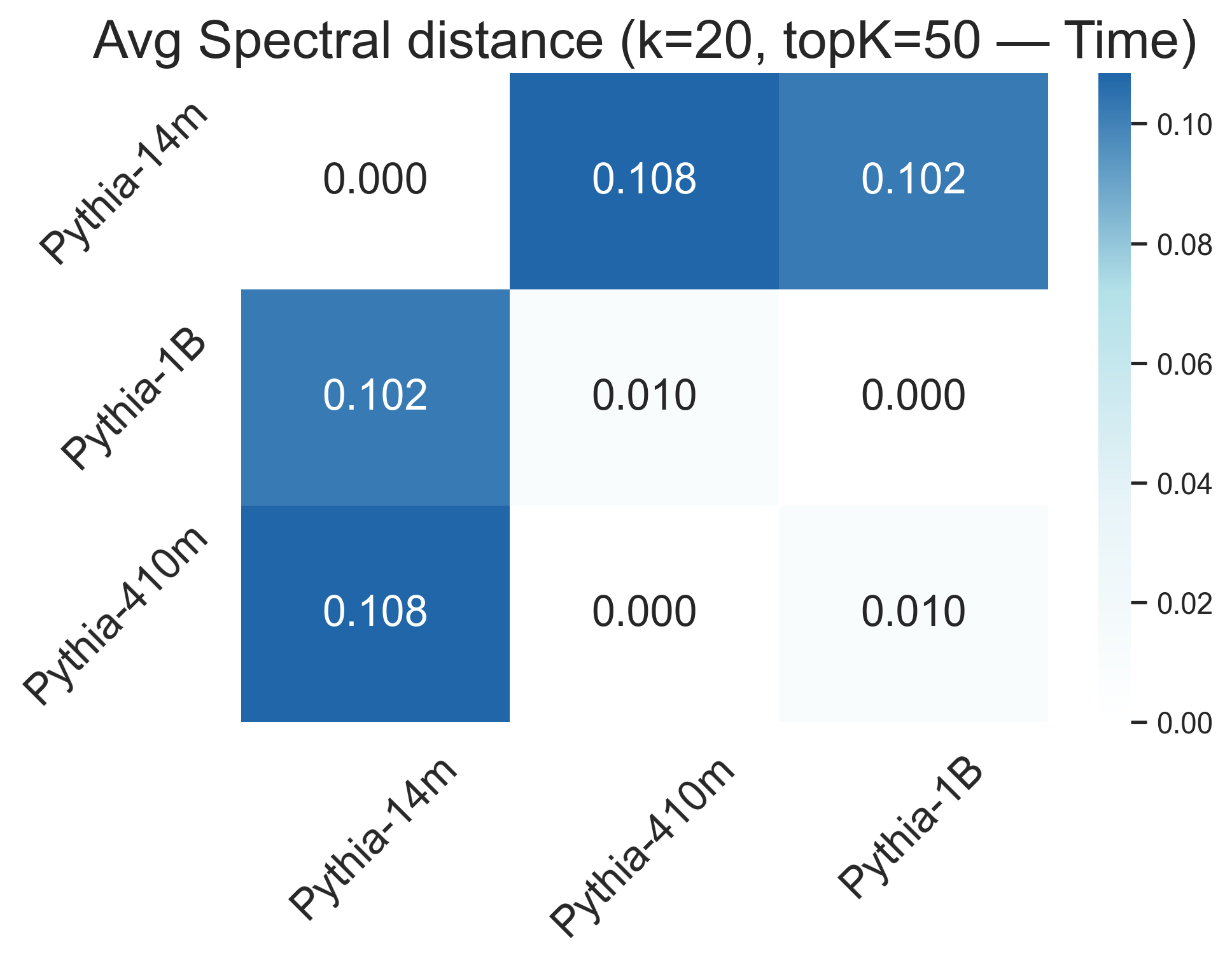}
    \caption{\textbf{Cross-scale spectral distance of role circuits.} Heatmaps show pairwise spectral distances between role circuits across \textsc{Pythia} model scales (14M, 410M, 1B), computed from the Top-50 edges and the first 20 eigenvalues. Lower values between larger models indicate greater similarity in information-flow geometry despite possible topological differences.}
    \label{fig:scale_spectral}
\end{figure}

\begin{figure*}
    \centering
    \includegraphics[width=0.8\linewidth]{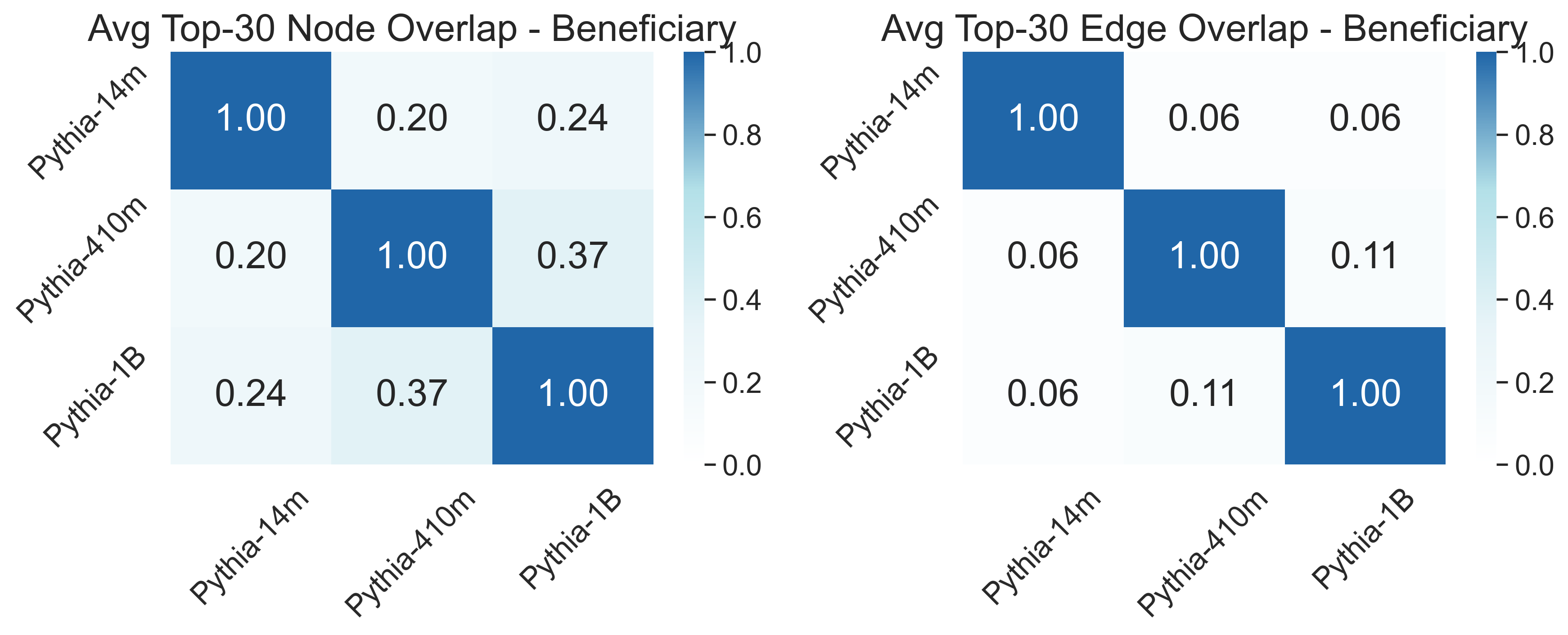}
    \includegraphics[width=0.8\linewidth]{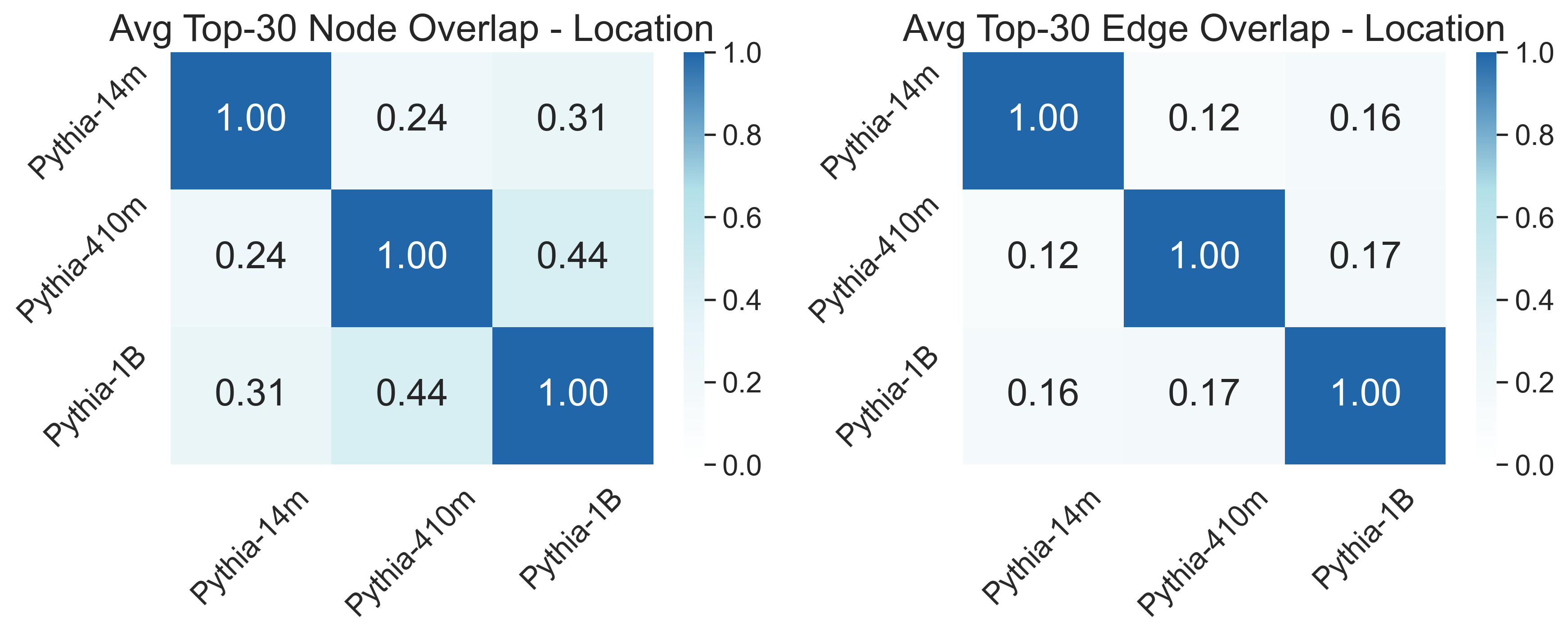}
    \includegraphics[width=0.8\linewidth]{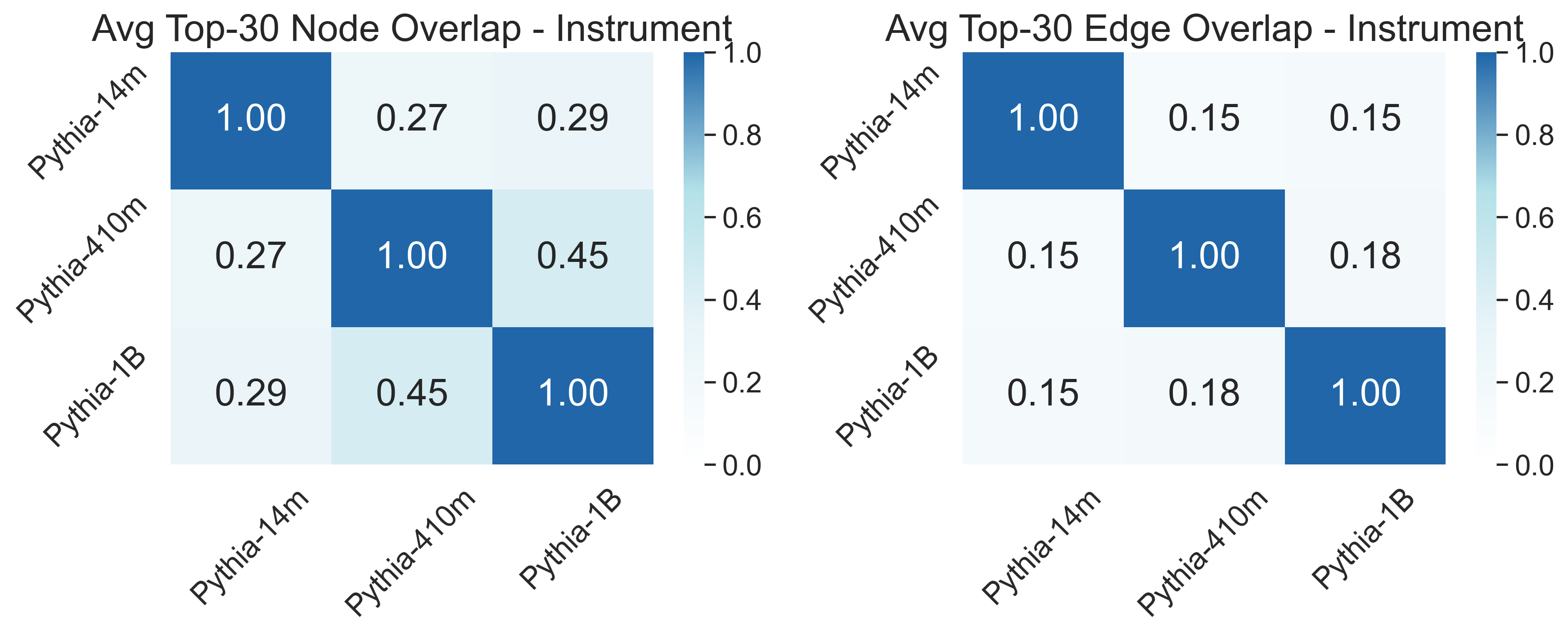}
    \includegraphics[width=0.8\linewidth]{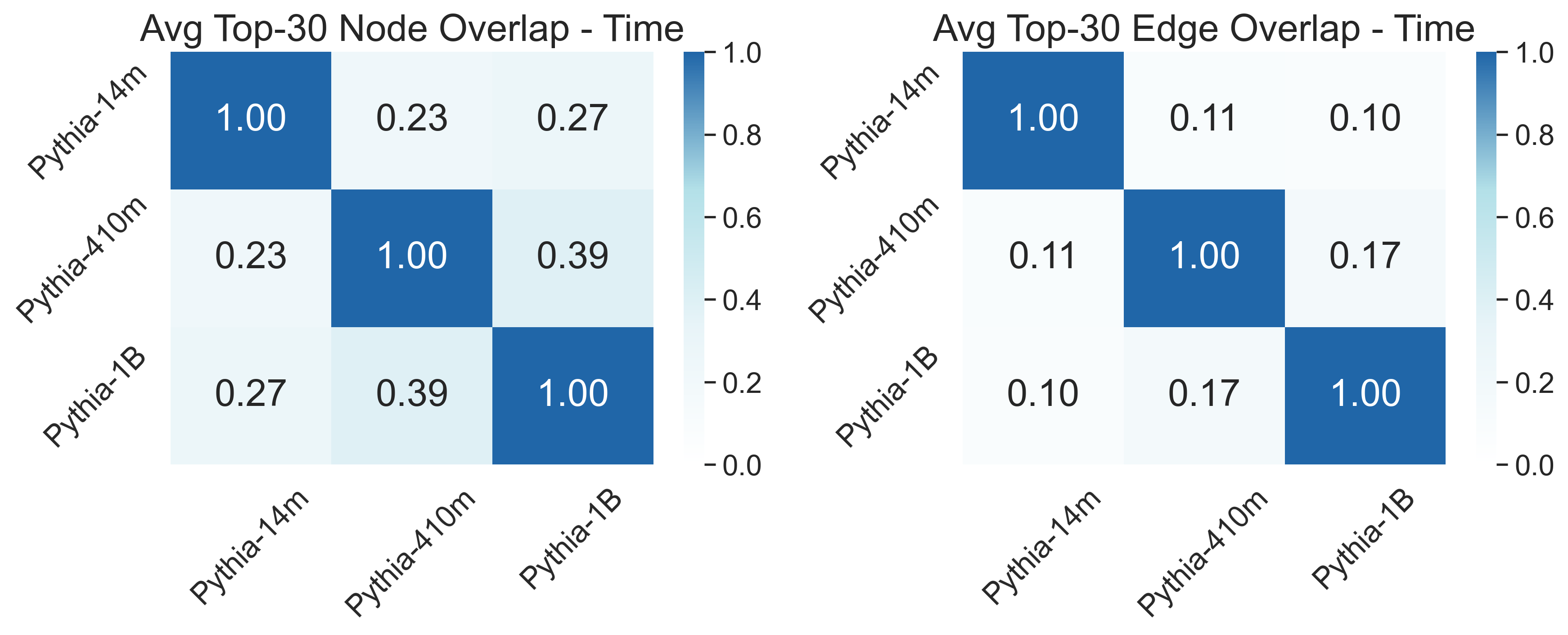}
    \caption{\textbf{Cross-scale overlap of high-importance components.} Heatmaps report average Top-30 node overlap (left) and Top-30 edge overlap (right) between role circuits across \textsc{Pythia} model scales. Node overlap consistently exceeds edge overlap, indicating reuse of key components with divergent routing across scales.}
    \label{fig:cross_scale_overlap}
\end{figure*}

\paragraph{Cross-scale similarity of component sets.}
Figures~\ref{fig:scale_spectral} and~ \ref{fig:cross_scale_overlap} summarise how role circuits align across \textsc{Pythia--14M}, \textsc{Pythia--410M}, and \textsc{Pythia--1B}. Node-level Top--30 overlaps vary by role, ranging from $J_V{\approx}0.24$--$0.31$ between \textsc{14M} and \textsc{1B} to $J_V{\approx}0.37$--$0.45$ between \textsc{410M} and \textsc{1B}. \textsc{Instrument} shows the strongest cross-scale alignment (up to $J_V{=}0.45$ for \textsc{410M}$\leftrightarrow$\textsc{1B}), indicating that its circuits are both highly compact and structurally similar at larger scales. Edge-level overlaps are consistently lower ($J_E{\approx}0.06$--$0.18$), confirming that connection patterns diverge more than the identity of high-importance nodes. Spectral distances are smallest for the \textsc{410M}$\leftrightarrow$\textsc{1B} pairs ($d_{\text{spec}}{\approx}0.006$--$0.018$) and larger when \textsc{14M} is involved ($d_{\text{spec}}{\approx}0.05$--$0.15$), suggesting that while small models already recruit broadly similar components, the overall information-flow geometry only stabilises once scale increases.

\paragraph{Summary.}
Across all roles and model scales, we find that semantic-role circuits localise to highly compact subgraphs whose attributional mass is dominated by a small, stable subset of nodes. Final-step Top--20 mass consistently exceeds 0.83 (and reaches 0.97--0.99 for several roles; Table~\ref{tab:rq1-sparsity-full}), and only $\sim$15--28 nodes are required to capture 95\% of total mass (Table~\ref{tab:rq1-coverage-full}). These component sets remain stable across training checkpoints, with only minor turnover in the highest-mass nodes. Structural metrics further reveal a characteristic pattern of refinement: active node sets contract slightly over training while density increases, indicating consolidation around a pruned but increasingly interconnected core. Together, these results show that role circuits are both \emph{spatially localised} and \emph{structurally coherent}, forming compact causal pathways that become progressively more organised as training proceeds.

\subsection{RQ2: Full Emergence Dynamics}\label{app:rq2-details}
This appendix complements the main-text analysis by providing the full emergence dynamics results for \textsc{Pythia–14M}. We analyse three signals across training: faithfulness, indispensability, and structural consolidation, and compare them to change-point estimates derived from piecewise linear fits.

\paragraph{Indispensability.}
All roles eventually become causally necessary, but the timings vary by more than four orders of magnitude. \textsc{Instrument} circuits are useful from the first checkpoint ($t_{\mathrm{ind}}{=}0$), \textsc{Location} becomes indispensable early (1k steps), and \textsc{Goal} follows at 5K. \textsc{Time} emerges only at mid-training ($71$k steps), reflecting exceptionally delayed functional reliance. These heterogeneous timings indicate that roles differ substantially in both cue learnability and the optimisation pressure required for the model to commit to a stable causal pathway.

\paragraph{Faithfulness trajectories.}
Faithfulness curves exhibit pronounced non-monotonicity, with early rises, sharp drops, and late partial recoveries. \textsc{Instrument} peaks early and declines; \textsc{Time} rises initially, crashes after 5–10k steps, and partially recovers; \textsc{Goal} and \textsc{Location} show smoother but still multi-phase dynamics. Importantly, these fluctuations do not correspond to abrupt structural changes: functional utility is unstable even when the underlying structure is already highly concentrated.

\paragraph{Structural consolidation.}
In contrast to faithfulness, structural metrics evolve smoothly. Top-$K$ node sets stabilise extremely early for most roles (512 steps for \textsc{Goal}, \textsc{Location}, and \textsc{Time}; 2K for \textsc{Instrument}). Thus, the model identifies the relevant components long before those components become functionally indispensable. Structural consolidation is therefore not the bottleneck in circuit emergence.

\paragraph{Change-point analysis.}
Two-segment piecewise fits to Top-$K$ mass trajectories yield very wide bootstrap confidence intervals (e.g.\ \textsc{Time}: [54, 5k]; \textsc{Instrument}: [64, 8K]), confirming that structural sparsification is gradual rather than concentrated at a discrete transition. Visual inspection of sparsity curves reveals smooth monotonic growth without identifiable inflection points. The structural substrate evolves continuously even when functional utility displays sharp changes.

\paragraph{Final-step sparsity.}
By convergence (143K steps), all circuits are highly compact: Top-20 mass ranges from 0.83–0.98, and only 15–18 nodes suffice to cover 95\% of attribution. \textsc{Beneficiary} is the most concentrated (95\% mass in 15 nodes), whereas \textsc{Instrument} and \textsc{Location} have slightly broader but still compact top-tiers.

\paragraph{Summary.}
Across all roles in \textsc{Pythia–14M}, emergence is a gradual process in which \emph{structural} properties stabilise early and monotonically, while \emph{functional} utility develops in a noisy, role-dependent manner. Indispensability can lag far behind consolidation, indicating that circuits may be structurally ``pre-allocated'' long before the model consistently relies on them. Together, these results support the conclusion that semantic-role circuits do not undergo discrete phase transitions but instead emerge through continuous refinement shaped by heterogeneous task signals and optimisation dynamics.

\subsection{Additional Semantic Roles}\label{app:other-roles}
\begin{figure*}[h!]
    \centering
    \includegraphics[width=\linewidth]{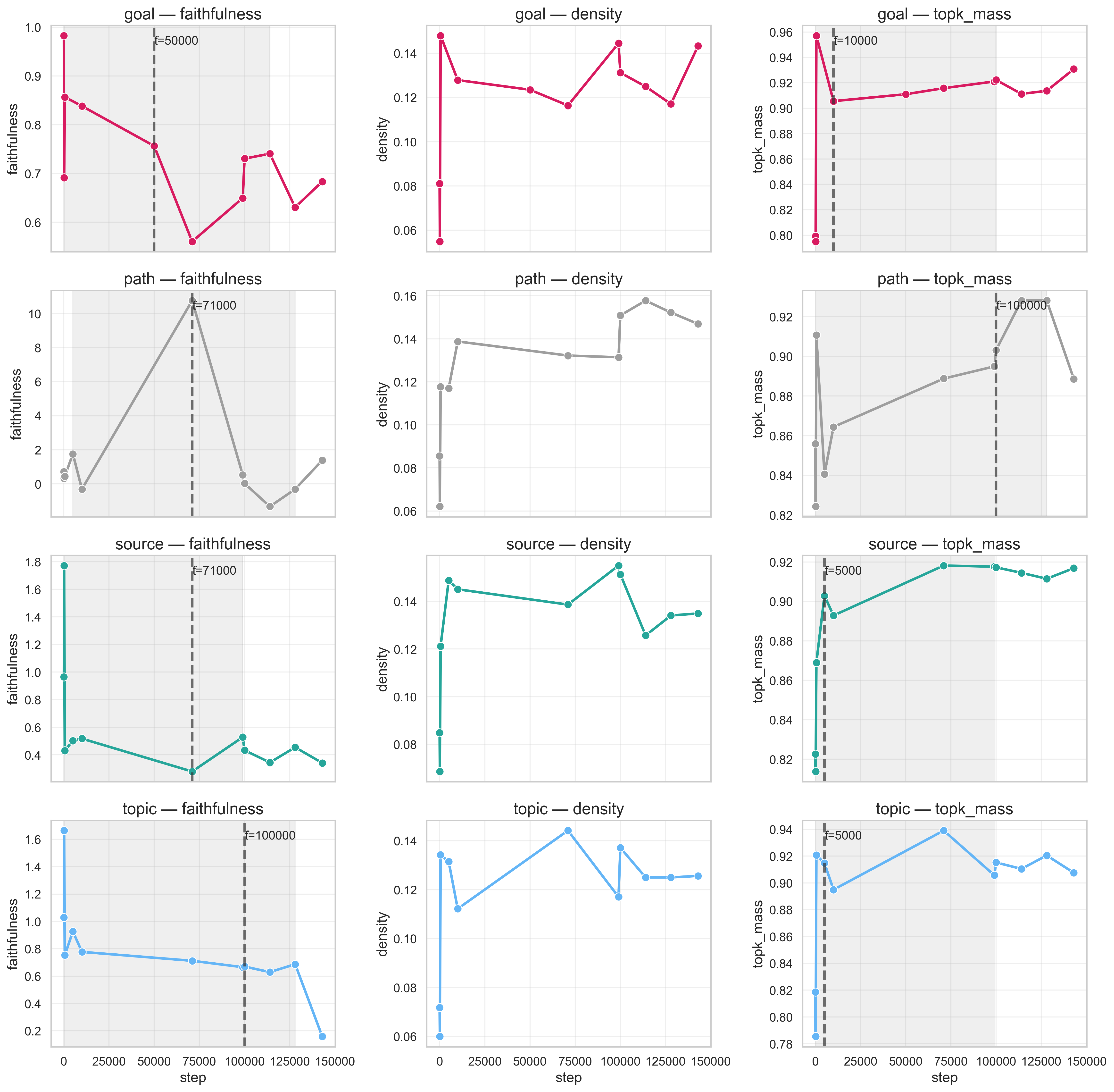}
    \caption{\textbf{Emergence trajectories for supplementary semantic roles for the \textsc{Pythia-1B} model.}
    Each row corresponds to a role; columns show faithfulness, density, and Top-$K$ mass across training.}
    \label{fig:other-roles-emergence}
\end{figure*}

In addition to the four core roles analysed in the main paper, we applied the COMPASS pipeline to four further predicate–argument relations frequently used in semantic role labelling, including  \textsc{Path}, \textsc{Source}, and \textsc{Topic}. Figure~\ref{fig:other-roles-emergence} reports their emergence
trajectories (faithfulness, density, Top-$K$ mass) across training checkpoints, and Tables~\ref{tab:rq2-tables-full-1b}, \ref{tab:rq2-sparsity-full-1b}, and \ref{tab:rq2-changepoint-topk-full-1B} demonstrate the indispensability, concentrations and change-point estimation for them, respectively. 

Overall, these supplementary roles exhibit comparable qualitative patterns identified for the main roles. Structural metrics (density and Top-$K$ mass) increase smoothly and monotonically, consistent with gradual sparsification rather than discrete phase transitions. 
In contrast, faithfulness trajectories are highly variable: \textsc{Path} exhibits high volatility: faithfulness increases through early training, then spikes mid-training (step 71k) before collapsing back, indicating heavy competition with other predictive cues. \textsc{Source} shows high faithfulness early training, the declines and remains consistently throughout training despite structural consolidation. \textsc{Topic} displays an early spike before settling into moderate stability. \textsc{Goal} maintains moderate faithfulness with fluctuations mid-training, never achieving the stability seen in others. These patterns match the dissociation seen in the primary roles: \emph{structural sparsity is stable and monotonic, but functional utility is noisy and task-pressure dependent}. As before, structure often stabilises long before a role becomes functionally useful.

\begin{figure*}[h!]
\centering
\includegraphics[width=\linewidth]{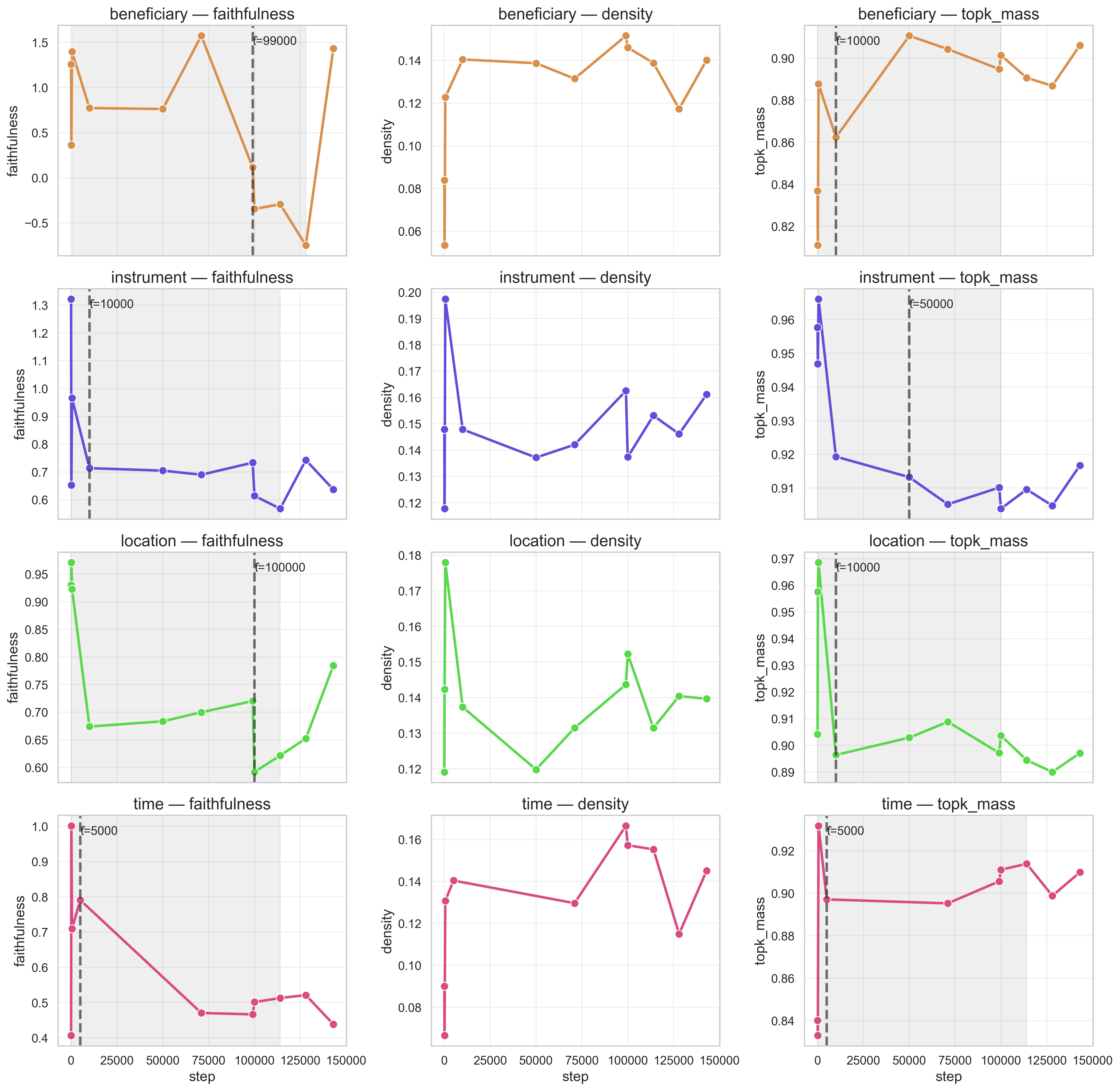}
\caption{Emergence trajectories for each role in \textsc{Pythia-1B}. Left: faithfulness; middle: edge density; right: Top-$K$ node mass (Top-20). Structural measures change smoothly over training, while faithfulness exhibits role-specific non-monotonicity.}
\label{fig:emergence-curves}
\end{figure*}

\begin{table}[t]
\centering
\small
\begin{tabular}{lcc}
\toprule
\textbf{Role} & $t_{\mathrm{ind}}$ (steps) & $t_{\mathrm{cons}}$ (steps) \\
\midrule
\textsc{Goal} & 0 & 1{,}000  \\
\textsc{Path} & 64 & 1{,}000  \\
\textsc{Source}  & 512 & 5{,}000 \\
\textsc{Topic} & 512 & 5{,}000 \\
\bottomrule
\end{tabular}
\caption{\textbf{Indispensability and consolidation timings for additional roles (\textsc{Pythia-1B} model).} The table reports the earliest step at which each role becomes indispensable ($t_{\mathrm{ind}}$), and the step at which structural consolidation occurs ($t_{\mathrm{cons}}$)}
\label{tab:rq2-tables-full-1b}
\end{table}

\begin{table}[t]
\centering
\small
\setlength{\tabcolsep}{3pt}
\begin{tabular*}{\columnwidth}{@{\extracolsep{\fill}}lcccc@{}}
\toprule
Role & Top-5 & Top-10 & Top-20 & $k$ for 80/90/95\% \\
\midrule
\textsc{Goal}  & 0.507&0.708 &\textbf{0.931} & 14 / 18 / 22 \\
\textsc{Path}  & 0.348 & 0.594 &\textbf{0.889} & 16 / 21 / 25 \\
\textsc{Source} & 0.421&0.655&\textbf{0.917} & 15 / 19 / 23 \\
\textsc{Topic}  & 0.444&0.657 &	\textbf{0.908} & 15 / 20 / 25 \\
\bottomrule
\end{tabular*}
\caption{\textbf{Final-step concentration of additional role circuits.} Similar to results in the main paper, a small $k$ values show a limited subset of nodes constituent the majority part of computation.}
\label{tab:rq2-sparsity-full-1b}
\end{table}

\begin{table}[t]
\centering
\small
\begin{tabular}{lcc}
\toprule
\textbf{Role} & $\hat{t}_c$ (steps) & 95\% CI \\
\midrule
\textsc{Goal} & 10{,}000  & [128, 10{,}000] \\
\textsc{Path} & 10{,}000 & [128, 128{,}000] \\
\textsc{Source}& 5{,}000   & [128, 99{,}000] \\
\textsc{Topic} & 5{,}000   & [128, 99{,}000] \\
\bottomrule
\end{tabular}
\caption{\textbf{Change-point estimates for Top-$K$ node mass.} Estimates are obtained via two-segment piecewise linear regression, with 95\% bootstrap confidence intervals shown for each role; wide intervals indicate gradual structural evolution.}
\label{tab:rq2-changepoint-topk-full-1B}
\end{table}

\subsection{Circuit heterogeneity}\label{sec:circuit_heterogeneity}
While our results demonstrate that semantic role circuits consistently localise to compact subgraphs, the \emph{internal organisation} of these circuits varies systematically across roles and training stages. This variation reflects both the diversity of semantic cues associated with different roles and the flexibility of the model's computational pathways. To characterise these differences, we examine the fine-grained structure of each circuit, its dominant components, routing patterns, and evolution over training, using causal-flow visualisations derived from the top-$K{=}30$ nodes, with edges ranked by attribution magnitude at the 95th percentile threshold. These analyses reveal systematic and role-dependent heterogeneity in circuit structure, both at convergence and throughout developmental trajectories.

\subsubsection{Architectural Stratification at Convergence}
We identify four recurrent circuit architectural types at convergence (step 143K), distinguished by attention-head addition, integration depth, and reliance on value composition operations. Table~\ref{tab:circuit_taxonomy_updated} summarises the final circuit types, reporting node counts, attention head involvement, and dominant computational pathways.

\begin{table}[h!]
\centering
\resizebox{\columnwidth}{!}{
\begin{tabular}{@{}lllcp{4.5cm}@{}}
\toprule
\textbf{Type} & \textbf{Role} & \textbf{Nodes} & \textbf{Heads} & \textbf{Architecture} \\
\midrule
1 & Path & 18 & 2 & Predominantly MLP with early framing (a0.h6) and late value composition (a12.h2+V) \\
  & Goal & 18 & 3 & Mid-layer extraction (a6.h0) with late integration (a12.h2+V, a14.h5) \\
\midrule
2 & Topic & 19 & 4 & Early frame detection (a0.h2), mid-layer (a6.h0), late integration (a12.h2, a14.h5+V) \\
  & Source & 19 & 5 & Rich multi-stage: early (a3.h7), mid (a6.h0), late (a12.h2, a14.h5, a15.h5) \\
\midrule
3 & Instrument & 20 & 4 & Mid-to-late extraction and integration (a3.h7, a6.h0, a12.h2, a14.h5) \\
  & Location & 20 & 4 & Distributed integration (a1.h0, a12.h2, a13.h4, a14.h5) \\
\midrule
4 & Beneficiary & 22 & 6 & Complex late-stage architecture (a1.h0, a6.h0, a12.h2, a14.h5+V, a15.h0, a15.h5) \\
& Time & 22 & 7 & Diverse multi-stage architecture with integration of several components (a3.h7, a3.h5, a6.h0, a9.h2, a10.h1, a12.h2+V, a14.h5+V, a15)\\
\bottomrule
\end{tabular}
}
\caption{\textbf{Circuit categorisation across semantic roles at convergence.} Circuits stratify into four architectural types reflecting the computational demands of each role. Node counts reflect circuits extracted at the 95th percentile threshold with top-$K{=}30$ edges. ``+V'' indicates value composition edges; heads are listed in order of prominence in the causal-flow diagrams.}\label{tab:circuit_taxonomy_updated}

\end{table}

\paragraph{Type 1: Lexical pattern matching.} \textsc{Goal} and \textsc{Path} rely predominantly on MLP computation with minimal or zero attention. \textsc{Path} exhibits minimal attention involvement (18 nodes, 2 heads), we hypothesise a0.h6 provides early syntactic composition, while a12.h2 performs targeted value composition for the final pre-logit refinement. The dominant causal flow remains through the MLP backbone, with attention providing what seems to be an auxiliary support. \textsc{Goal} (18 nodes, 3 heads) implements a clean two-stage architecture: mid-layer feature extraction at a6.h0, followed by late-stage integration at a12.h2+V (with value composition edge) and final refinement at a14.h5. The circuit exhibits sparse connectivity with clear information bottlenecks at m6 and m12. Both roles mark a highly regular, closed set of expressions (e.g., path markers), enabling direct lexical classification without extensive need for contextual integration.

\paragraph{Type 2: Multi-stage compositional integration.} \textsc{Topic}, and \textsc{Source} converge to moderate-complexity architectures (19 nodes, 4--5 attention heads) characterised by systematic multi-stage processing, with m0 being connected to the majority of other nodes in the circuit.  \textsc{Topic} (19 nodes, 4 heads) adds early integration of a0.h2, suggesting initial scaffold detection, then follows mid-to-late integration (a6.h0$\rightarrow$a12.h2$\rightarrow$a14.h5+V). Notably, value composition shifts to the final integration head (a14.h5+V) rather than at a12.h2, indicating part of feature extraction occurs at the pre-logit stage. \textsc{Source} (19 nodes, 5 heads) exhibits the richest attention architecture in this group: a3.h7, assumably for early instrumental/causal cue detection, a6.h0 for mid-layer feature gathering, and three late-stage heads (a12.h2, a14.h5, a15.h5) for final disambiguation. The circuit displays clear information flow through mid-layer MLPs (m3, m6) converging to late-stage integration at m12 and m15, with a15.h5 providing a final refinement pathway.

\paragraph{Type 3: Balanced hybrid architectures.} \textsc{Instrument} and \textsc{Location} exhibit moderate complexity (20 nodes, 4 heads each) with balanced MLP-attention integration. \textsc{Instrument} follows a hierarchical pattern: a3.h7 for early cue detection (likely detecting instrumental markers), a6.h0 for mid-layer extraction, and late-stage integration at a12.h2 and a14.h5. The circuit exhibits a clear bottleneck at m6, with multiple pathways converging to m12 and m14. Notably, \textsc{Instrument} lacks value composition at convergence, relying instead on positional routing through residual connections. \textsc{Location} shows a more distributed architecture with attention heads spread across early (a1.h0), mid-late (a12.h2, a13.h4), and final (a14.h5) layers. The circuit maintains sparse MLP connectivity (m1, m13, m14) with attention providing targeted integration at multiple depths. Like \textsc{Instrument}, \textsc{Location} lacks value composition edges, suggesting both roles have do not rely as much as others on active feature extraction mechanisms in favour of simpler, more efficient routing strategies by convergence.

\paragraph{Type 4: Complex late-stage integration with distributed refinement.} \textsc{Beneficiary} and \textsc{Time} exhibit the most elaborate architectures at convergence (22 nodes each, 6--7 attention heads), maintaining rich late-stage connectivity with distributed processing across multiple layers. The circuit recruits a1.h0 for early processing, a6.h0 for mid-layer extraction, and four late-stage heads: a12.h2 for primary integration, a14.h5+V for feature extraction, and two heads (a15.h0, a15.h5) for final disambiguation. This architecture suggests \textsc{Beneficiary} requires parallel processing pathways to resolve persistent ambiguities, and probably the model maintains alternative hypotheses until the last computation step. The circuit's relative complexity (22 nodes vs. 16--20 for other roles) and late-stage density indicate that beneficiary marking, despite being syntactically constrained (e.g., ``for $\dots$''), requires more elaborate compositional reasoning than other participant roles, likely to distinguish benefactive readings from alternative interpretations. \textsc{Time} (22 nodes, 7 heads) exhibits similarly complex multi-stage processing with attention distributed across early (a3.h7), mid-layer (a6.h0, a9.h2, a10.h1), and late integration (a12.h2, a14.h5+V, a15). The circuit shows convergent information flow through mid-layer MLPs (m6, m9, m10) to late-stage integration nodes (m12, m14), with value composition at a14.h5 enabling feature extraction for temporal disambiguation. This architectural complexity is surprising given that temporal expressions often involve closed-class markers (``during the'', ``at the''), but likely reflects the need to distinguish temporal from locative interpretations of ambiguous scaffolds (``at the'') and to resolve context-dependent temporal reference. 

\subsubsection{Developmental Trajectories Across Training}

\begin{figure*}[ht]
    \centering
        \includegraphics[width=0.6\textwidth]{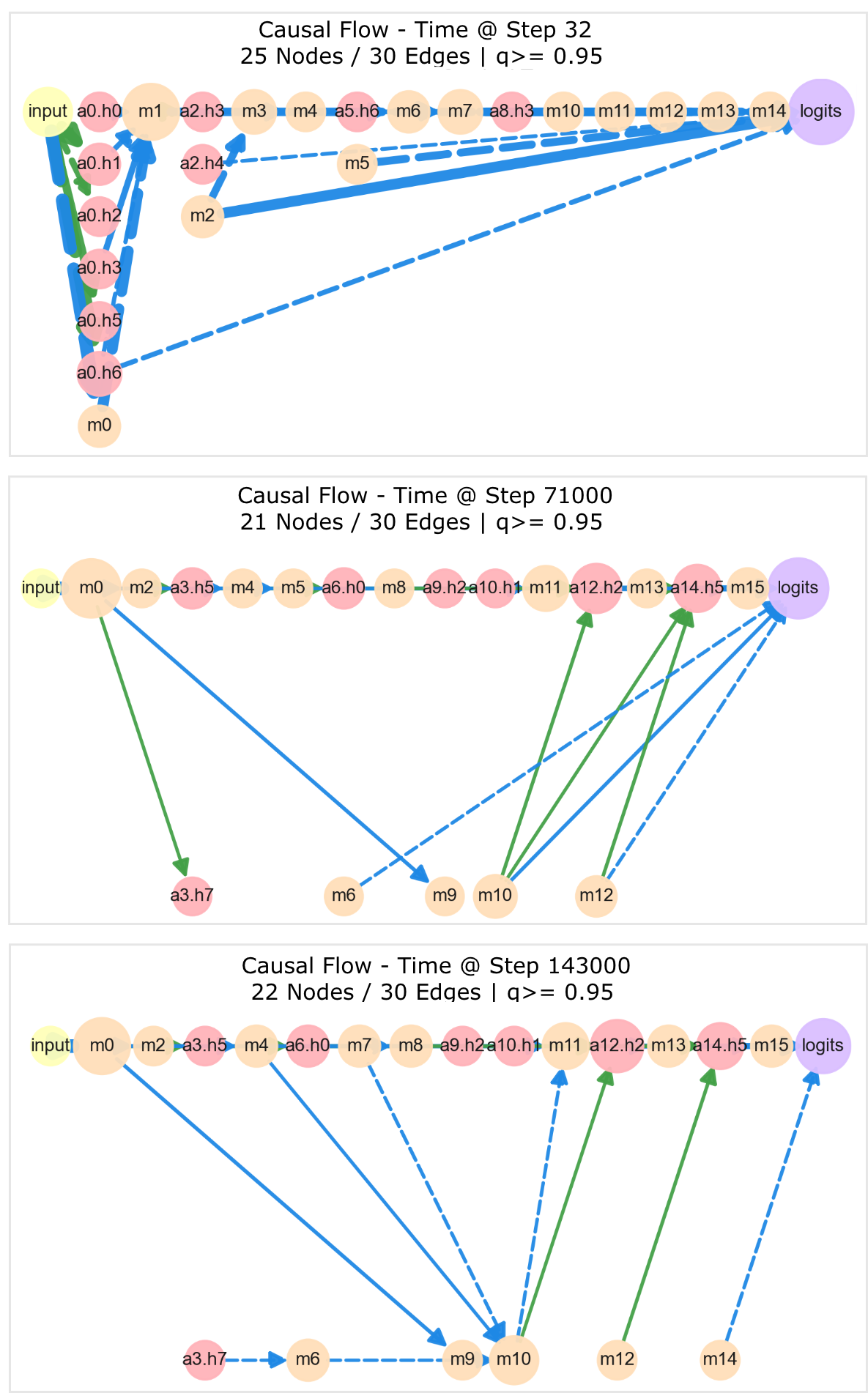}
    \caption{\textbf{Developmental trajectory for \textsc{Time} role across training.} Circuit evolution from initialisation (step 32) through mid-training (step 71000) to convergence (step 143000) demonstrates progressive attention elimination.}
    \label{fig:time_evolution}
\end{figure*}

\begin{figure*}[ht]
    \centering
        \includegraphics[width=0.6\textwidth]{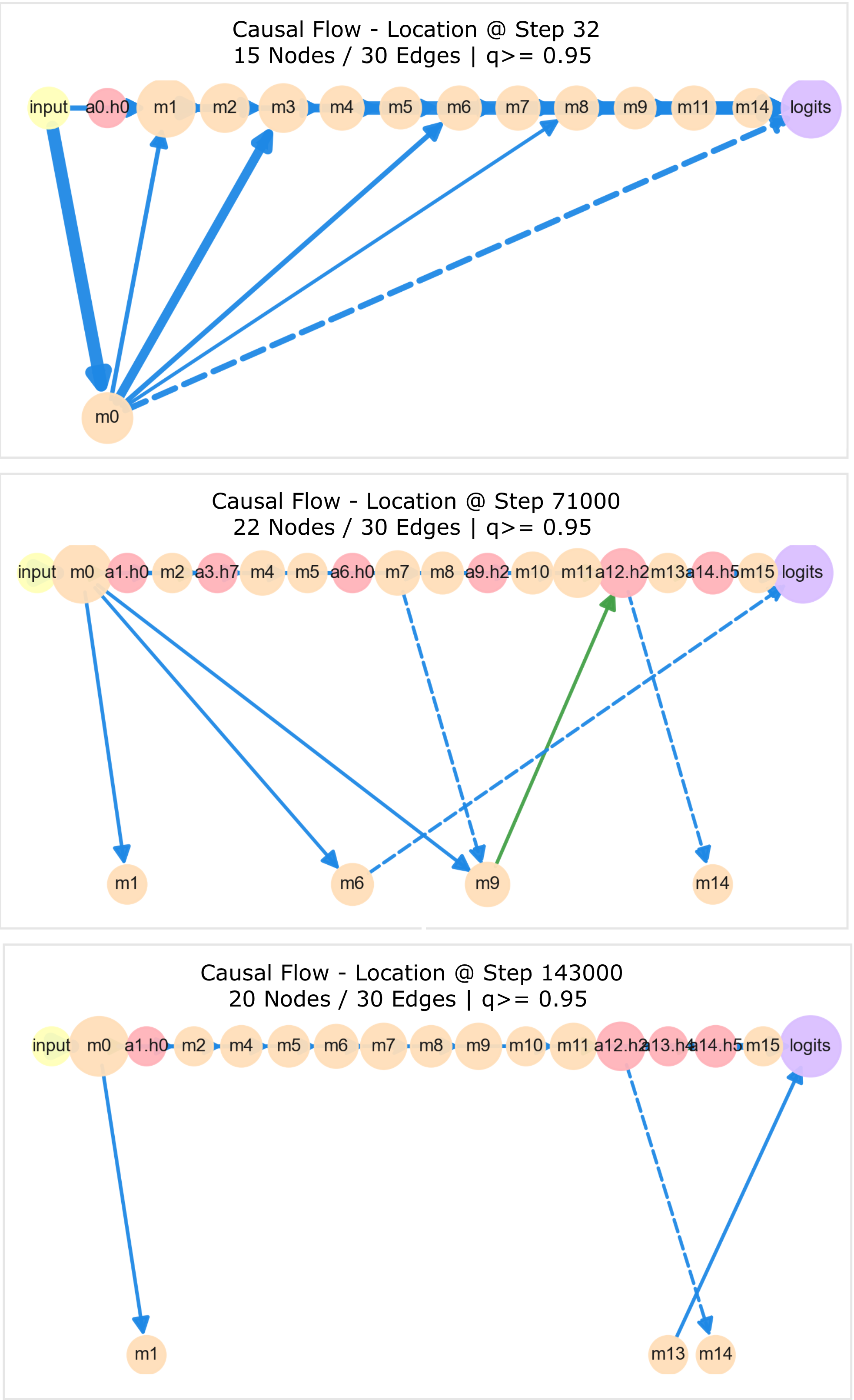}
        \caption{\textbf{Developmental trajectory for \textsc{Location} role across training.} Circuit evolution follows an expand then contract pattern with mid-training exploration followed by late-stage pruning.}
    \label{fig:location_evolution}
\end{figure*}

\begin{figure*}[ht]
    \centering
        \includegraphics[width=0.6\textwidth]{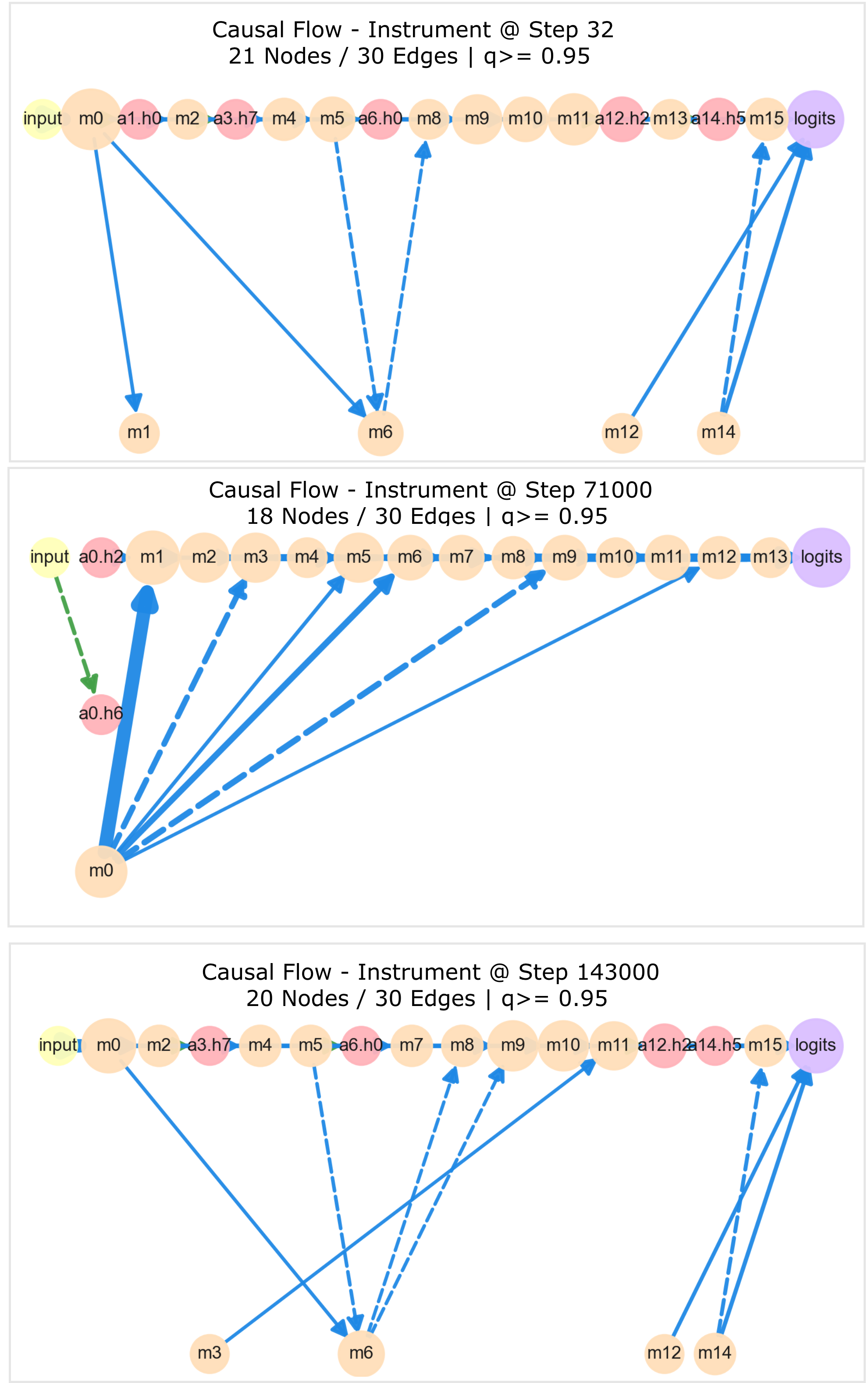}
        \caption{\textbf{Developmental trajectory for \textsc{Instrument} role across training.} Circuit evolution demonstrates stable mid-layer consolidation with minimal late-stage refinement.}
    \label{fig:instrument_evolution}
\end{figure*}

Circuit evolution from initialisation to convergence (step 143K) reveals role-specific developmental patterns. Figures~\ref{fig:time_evolution}, \ref{fig:location_evolution} and \ref{fig:instrument_evolution} present a sample of causal-flow diagrams at three key checkpoints, early training (step 32), mid-training (step 71000), and convergence (step 143000). These snapshots expose systematic differences in how semantic-role circuits emerge and stabilise. We study them for all roles below. 

\paragraph{Universal early complexity followed by selective pruning.}
All roles begin training with diffuse connectivity and elevated node counts (15--23 nodes at step 32), reflecting initial hypothesis exploration. At initialisation, most circuits exhibit dense early-layer attention recruitment (multiple a0.h* heads) and extensive value composition edges, suggesting the model initially explores compositional integration strategies broadly across all roles. By mid-training (step 71000), circuits have begun to differentiate sharply: \textsc{Time} achieves near-complete attention elimination (16 nodes, pure MLP), while \textsc{Beneficiary} maintains a rich multi-head architecture (21 nodes, 7 heads). At convergence (step 143000), final node counts (16--22) represent reductions from initialisation, with architectural consolidation complete and further refinement involving only edge-weight adjustments rather than topological restructuring.

\paragraph{Stratified consolidation and shared infrastructure.}
Roles follow distinct consolidation trajectories aligned with semantic complexity. What we assume are more \textbf{Template-matching roles} (\textsc{Goal}, \textsc{Path}) eliminate most attention: \textsc{Path} retains minimal heads (a0.h6, a12.h2+V). \textbf{Integration-dependent roles} ( \textsc{Instrument}, \textsc{Location}) stabilise core architectures by mid-training with only minor late refinement, converging to 2--4 attention heads with mid-to-late integration. \textbf{Multi-phase processing roles} (\textsc{Topic}, \textsc{Source}, \textsc{Beneficiary}) undergo non-monotonic reorganisation: \textsc{Topic} contracts then reinstates early framing (a0.h2) and late value composition (a14.h5+V); \textsc{Source} adds a3.h7 only at convergence; \textsc{Beneficiary} and \textsc{Time} maintain persistent complexity with many heads likely required for parallel disambiguation.

Across trajectories, shared integration hubs emerge at predictable stages: {a12.h2} (7/8 roles) stabilises by mid-training as universal late-stage integrator; {a14.h5} (7/8 roles) emerges slightly later for pre-logit refinement; {a6.h0} (5/8 roles) provides mid-layer extraction for compositional/discourse roles. Early-layer attention (a0.h*, a1.h*) is systematically pruned except in \textsc{Path}, \textsc{Topic}, \textsc{Location}, and \textsc{Beneficiary}, where explicit frame detection remains necessary. Value composition follows more of explore/prune/retain dynamics: broad at initialisation (6/8 roles), divergent at mid-training (4 roles eliminate, 1 intensifies to 3 V-ops), and selective at convergence (4 roles: \textsc{Path}, \textsc{Goal}, \textsc{Topic}, \textsc{Beneficiary}). V-edges shift spatially from early layers (a0.h*, a3.h*) to late integration heads (a12.h2, a14.h5), indicating feature extraction is preserved only where disambiguation proves irreducible.

\subsection{Paraphrase-Based Within-Role Scaffold Controls (Pythia–1B)}
\label{app:paraphrase-controls}

To test whether role circuits encode \emph{abstract semantic roles} rather than overfitting to a particular lexicalised scaffold, we construct a small within-role paraphrase control set for two representative roles: \textsc{Location} and \textsc{Instrument}, and we study them over our reference model \textsc{Pythia–1B}. 



\paragraph{Paraphrase construction.}
For each filtered example $(x^{(r)}, y^{(r)})$ in the role-cross dataset (Appendix~\ref{app:data}), we construct a within-role paraphrase by replacing the original scaffold with an alternative:

\begin{equation}
x^{(r)}_{\text{para}} = \text{``The agent verb theme scaffold}^{(r')}\text{''}
\end{equation}
by replacing $\mathrm{scaffold}^{(r)}$ with an alternative scaffold $\mathrm{scaffold}^{(r')} \in \mathcal{S}^{(r)}$ such that:
\begin{itemize}[leftmargin=*, itemsep=1pt]
    \item $\mathrm{scaffold}^{(r')} \neq \mathrm{scaffold}^{(r)}$
    \item $|\mathrm{toks}(\mathrm{scaffold}^{(r')})| = |\mathrm{toks}(\mathrm{scaffold}^{(r)})|$ (parity constraint)
    \item agent, verb, theme, and target head $y^{(r)}$ are unchanged
\end{itemize}
For example, 
\begin{align*}
\mathcal{S}^{(\textsc{Location})} 
    &= \{\text{``in the''},\ \text{``at the''},\ \text{``near the''}\}.\\
\mathcal{S}^{(\textsc{Instrument})} 
    &= \{\text{``with the''},\ \text{``using the''},\ \text{``by the''}\}.
\end{align*}
We then apply the same filtering criterion as for the main dataset (Section~\ref{app:data}): we retain only paraphrased prompts $x^{(r)}_{\text{para}}$ for which the model continues to predict the original role-appropriate target $y^{(r)}$ as the most probable next token. This yields a paraphrase set
\begin{equation}
\mathcal{D}_{\text{para}}^{(r)} = \{(x^{(r)}_{\text{para}}, y^{(r)})\}
\end{equation}
in which all examples are (i) semantically equivalent to the originals at the level of role assignment, (ii) realised with different surface scaffolds, and (iii) correctly handled by the base model.

\paragraph{Circuit consistency evaluation.}
For each role $r \in \{\textsc{Location},\ \textsc{Instrument}\}$, we recompute EAP-IG scores on the paraphrase set and compare them with the original role-cross circuit $\mathcal{C}^{(r)}$ using:
\begin{itemize}[leftmargin=*, itemsep=2pt]
    \item Top-$K$ node overlap ($K=20$),
    \item Edge-weight rank correlation,
    \item Faithfulness of original circuits on paraphrased prompts.
\end{itemize}

\paragraph{Summary of control results.}
Table~\ref{tab:paraphrase-results} summarises the results for \textsc{Instrument} and \textsc{Location}. Across both roles, we observe:


\begin{table}[h]
\centering


\resizebox{\columnwidth}{!}{
\begin{tabular}{lcccc}
\toprule
\textbf{Metric} & \textbf{n} & \textbf{Pearson $r$} & \textbf{p-value} & \textbf{Spearman $\rho$} \\
\midrule
Circuit Size (Nodes)      & 8 & -0.066 & 0.876 & 0.049 \\
Attention Heads           & 8 & -0.283 & 0.497 & -0.268 \\
Sparsity (Top-20 Mass)    & 8 &  0.296 & 0.476 & 0.311 \\
Concentration (Gini)      & 8 &  0.326 & 0.430 & 0.357 \\
Consolidation Time ($t_{cons}$) & 8 & -0.054 & 0.899 & 0.062 \\
\bottomrule
\end{tabular}
}
\caption{Correlations between role frequency (sample size) and circuit properties at convergence (Pythia-1B, step 143K). All correlations show, indicating weak relationships between role frequency and circuit properties. Sample sizes range from 491 to 1,212 examples per role.}
\label{tab:frequency_correlations}
\end{table}

\begin{table}[h!]
\centering
\small
\begin{tabular}{lcc}
\toprule
\textbf{Role} & \textbf{Faithfulness} & \textbf{Top--20 Mass} \\
\midrule
\textsc{Instrument}  & $0.657\;(+0.021)$ & $0.925\;(+0.009)$ \\
\textsc{Location}    & $0.731\;(-0.053)$ & $0.920\;(+0.023)$ \\
\bottomrule
\end{tabular}
\caption{Paraphrase-control evaluation for two roles in \textsc{Pythia--1B}.
Scores reported for paraphrased inputs at the final training step.  
Parentheses denote the change relative to the original (non-paraphrased) dataset.}\label{tab:paraphrase-results}
\end{table}
\begin{table}[h]
\centering
\resizebox{\columnwidth}{!}{
\begin{tabular}{lcccc}
\toprule
\textbf{Circuit Property} & \textbf{n} & \textbf{Pearson r} & \textbf{p-value} & \textbf{Spearman $\rho$} \\
\midrule
Circuit Size (nodes)      & 8 &  0.346 & 0.401 &  0.439 \\
Attention Heads           & 8 &  0.310 & 0.455 &  0.244 \\
Sparsity (Top-20 mass)    & 8 &  0.106 & 0.803 &  0.095 \\
Concentration (Gini)      & 8 &  0.430 & 0.288 &  0.381 \\
\bottomrule
\end{tabular}
}
\caption{Corpus frequencies from 5,000 documents sampled from The Pile and the correlations between corpus scaffold frequency and circuit properties at convergence (Pythia-1B, step 143K). All correlations show $|r| \leq 0.430$, $p \ge 0.288$ (not significant). }
\label{tab:corpus_freq_circuit_correlations}
\end{table}
\begin{table*}[ht]
\centering
\resizebox{\textwidth}{!}{
\begin{tabular}{lccccc}
\toprule
\textbf{Method} & \textbf{Faithful} & \textbf{Path-spec.} & \textbf{Granularity} & \textbf{Scalable} & \textbf{Notes / Risks}\\
\midrule
Linear probing & \(\circ\) & \(\times\) & token/residual & \(\checkmark\) & Correlational, not causal \\
Attribution Patching (EAP) \citep{nanda2023attribution, syed-etal-2024-attribution} & \(\circ\) & \(\checkmark\) & head/MLP edge & \(\checkmark\) & Gradient-based; false negatives \\
AtP$^\ast$ \citep{kramar2024atp} & \(\checkmark\) & \(\checkmark\) & head/MLP edge & \(\checkmark\checkmark\) & improved faithfulness; residual false negatives \\
Temporal EAP\text{-}IG* & \(\checkmark\) & \(\checkmark\) & head/MLP edge & \(\checkmark\checkmark\) & Baseline/path sensitivity \\
Path patching \citep{GoldowskyDill2023LocalizingMB} & \(\checkmark\) & \(\checkmark\checkmark\) & head/MLP path & \(\circ\) & Expensive over checkpoints \\
Causal scrubbing \citep{chan2022causal} & \(\checkmark\checkmark\) & \(\checkmark\checkmark\) & hypothesis-level & \(\times\) & High implementation cost \\
SAE \citep{templeton2024scaling} & \(\checkmark\) & \(\checkmark\) & feature-level & \(\circ\) & SAE training; feature drift \\
Transcoders \citep{dunefsky2024transcoders} & \(\checkmark\) & \(\checkmark\) & MLP sublayer & \(\circ\) & Surrogate training; not causal \\
Circuit tracing / attribution graphs \citep{ameisen2025circuit} & \(\circ\) & \(\checkmark\checkmark\) & feature--feature & \(\times\) & Surrogate fidelity; prompt-specific \\
\bottomrule
\end{tabular}
}
\caption{Comparison of interpretability methods along criteria induced by RQ1/RQ2. \(\checkmark\checkmark\)=strong; \(\checkmark\)=good; \(\circ\)=partial; \(\times\)=weak.}
\label{tab:method-comparison}
\end{table*}

\begin{enumerate}[leftmargin=*, itemsep=2pt]
    \item \textbf{Stable faithfulness under paraphrase}: the original circuits maintain similar causal impact despite scaffold changes (Instrument: $+0.02$; Location: $-0.05$).
    \item \textbf{Consistent sparsity structure}: Top--20 mass shifts by less than $0.03$ for both roles, indicating that high-importance components remain largely unchanged.
    \item \textbf{Robust abstraction beyond surface cues}: circuits respond similarly across distinct paraphrastic templates, supporting the interpretation that they encode predicate–argument binding rather than memorised lexical patterns.
\end{enumerate}

These findings reinforce that the circuits identified by our study capture the semantic structure associated with predicate–argument roles, rather than superficial or memorised prepositional cues.

\subsection{Role Frequency and Circuit Properties}\label{app:role_frequency_analysis}
\subsubsection{Sample Size and Circuit Properties}
\label{app:sample_size_circuits}
To test whether circuit architecture is determined by the availability of samples in our filtered dataset, we examined correlations between sample sizes and circuit properties at convergence (Pythia-1B, step 143K). If circuit complexity were driven primarily by the volume of valid examples, we would expect positive correlations between sample sizes and measures of circuit size or structural complexity.

\paragraph{Correlation Analysis.}
Table~\ref{tab:frequency_correlations} reports Pearson and Spearman correlations between filtered sample sizes and five circuit properties: circuit size (nodes), attention head count, sparsity (Top-20 mass), concentration (Gini coefficient), and consolidation timing ($t_{cons}$). All correlations show $|r| < 0.35$ with all $p > 0.4$, indicating negligible relationships between sample size and circuit architecture. The strongest correlation (Gini: $r = 0.326$, $p = 0.430$) remains statistically

\paragraph{Role-Type Stratification.}
In contrast to the absence of sample-size effects, circuit architecture stratifies clearly by role type (Table~\ref{tab:circuit_taxonomy_updated}). Roles requiring complex disambiguation (Type 4: BENEFICIARY, TIME) develop the largest circuits (22 nodes, 6--7 attention heads) independent of their sample sizes (621 vs. 753 examples). Conversely, lexical pattern matching (Type 1: PATH) converges to a minimal architecture (18 nodes, 2 heads) despite moderate sample availability (707 examples). Multi-stage compositional roles (Type 2) and balanced hybrid architectures (Type 3) occupy intermediate positions, with circuit complexity determined by semantic processing demands rather than example count.

The consistent absence of correlations between sample sizes and all circuit properties demonstrates that architectural complexity is not determined by the number of valid examples in our filtered dataset. 

\subsubsection{Scaffold Frequency in Training Corpus}
\label{app:scaffold_freq}
To validate that the selected semantic roles are well-represented in the training data and to test whether circuit architecture can be explained by training signal strength alone, we analysed scaffold frequencies in the Pythia training corpus (The Pile \citep{pile}). We sampled 5000 documents ($\sim$50M tokens) using stratified sampling (every 4,200th document) to ensure representative coverage across Pile subsets, and counted occurrences of the prepositional scaffolds used in our role-conditioned continuation task. We then correlated these corpus frequencies with circuit properties at convergence to test whether training frequency determines architectural complexity.

\paragraph{Corpus Frequencies.}
All roles are well-represented in the training data, ranging from 517.9 (\textsc{Topic}) to 5,498.9 (\textsc{Location}) instances per million tokens. Locative and temporal roles exhibit the highest frequencies (\textsc{Location}: 5,498.9/1M; \textsc{Time}: 5,452.3/1M), reflecting the prevalence of spatial and temporal modification in natural language. Directional roles show intermediate frequencies (\textsc{Goal}: 4,147.6/1M;  \textsc{Source}: 1,057.2/1M; PATH: 598.2/1M), while participant and propositional roles appear less frequently but remain well represented (\textsc{Instrument}: 1,990.5/1M; \textsc{Beneficiary}: 1,523.3/1M; TOPIC: 517.9/1M).

\paragraph{Correlation with Circuit Architecture.}
Table~\ref{tab:corpus_freq_circuit_correlations} reports Pearson and Spearman correlations between corpus scaffold frequencies and circuit properties at convergence (step 143K). All correlations are weak to moderate ($|r| \leq 0.430$) and statistically non-significant ($p > 0.28$), indicating that training frequency does not reliably predict circuit architecture. The strongest correlation (Gini coefficient: $r = 0.430$, $p = 0.288$) explains only 18\% of variance and is not statistically significant. Circuit size and attention head count show weak positive trends ($r = 0.346$ and $r = 0.310$ respectively), while sparsity (Top-20 mass) shows virtually no relationship ($r = 0.106$, $p = 0.803$). We note that one of these relationships approaches statistical significance at $\alpha = 0.05$ level.

The absence of systematic frequency effects is most clearly demonstrated by roles with nearly identical corpus frequencies but divergent architectures. \textsc{Location} and \textsc{Time} appear at virtually the same frequency in the training data (5,498.9 vs. 5,452.3 per million tokens, a difference of 0.8\%), but develop different circuit structures. \textsc{Location} converges to a balanced hybrid architecture (20 nodes, 4 attention heads) with distributed integration across layers (a1.h0, a12.h2, a13.h4, a14.h5), while \textsc{Time} develops a more complex multi-stage architecture (22 nodes, 7 attention heads) with rich attention involvement distributed across early (a3.h7), mid-layer (a6.h0, a9.h2, a10.h1), and late integration stages (a12.h2, a14.h5, a15). This architectural difference, despite a near-identical training signal, demonstrates that circuit organisation cannot be predicted from corpus frequency alone. The results indicate that instead of building circuits proportional to how often each role appears in training, models develop architectures tailored to the semantic processing demands of each role. This suggests that circuit organisation reflects functional requirements, the computational work needed to bind each semantic role, rather than statistical regularities in the training corpus.

\section{Method Comparison} \label{app:method-comparison}
We provide a summary comparing mechanistic interpretability methods, along with our choice in Table~\ref{tab:method-comparison}.

\end{document}